%% file: main.tex
\documentclass[10pt,twocolumn,letterpaper]{article}

\usepackage[pagenumbers]{cvpr} 

\input{preamble}

\definecolor{cvprblue}{rgb}{0.21,0.49,0.74}
\usepackage[pagebackref,breaklinks,colorlinks,citecolor=cvprblue]{hyperref}

\def\paperID{5987} 
\def\confName{CVPR}
\def\confYear{2024}

\graphicspath{ {./media/} }
\makeatletter
\def\blfootnote{\xdef\@thefnmark{}\@footnotetext}
\makeatother

\begin{document}

\title{``\emph{Previously on ...}''
From Recaps to Story Summarization}

\author{
Aditya Kumar Singh \hspace{1cm}
Dhruv Srivastava \hspace{1cm}
Makarand Tapaswi \\
CVIT, IIIT Hyderabad, India \\
{\small\url{https://katha-ai.github.io/projects/recap-story-summ/}}
}

\twocolumn[{%
\renewcommand\twocolumn[1][]{#1}%
\maketitle

\centering
\vspace{-4mm}
\includegraphics[width=1.0\linewidth]{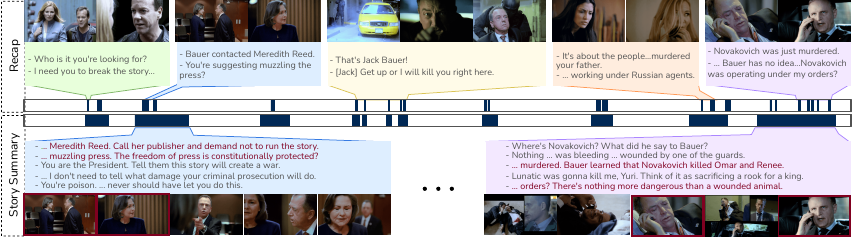}
\vspace{-6mm}
\captionof{figure}{
We illustrate how TV show recaps can be used to generate labels for \emph{multimodal story summarization}.
The \emph{top half} features the recap shown at the beginning of the episode S08E23 based on key moments (shots and dialogs) from S08E22 of the TV series \emph{24}. 
As recaps help viewers recall essential story events, we extend these aligned segments to create summarization labels (visualized in the \emph{bottom half} where the actual shots and dialogs inherited from recap are marked in {\color{darkred} deep red}).
For example, in the {\color{royalblue} sub-story} (left), the recap hints at Jack Bauer relaying classified information to the press, while the summary presents the complete sub-story, including Logan informing President Taylor about their failure to catch Jack and their disagreement over muzzling the press.
}
\label{fig:teaser}
\vspace{8mm}
}]


\input{sections/00_abstract}

\input{sections/01_introduction}
\input{sections/02_related_works}

\input{sections/03_dataset}

\input{sections/04_method}

\input{sections/05_experiments}

\input{sections/06_conclusion}


\clearpage
{
\small
\bibliographystyle{ieeenat_fullname}
\bibliography{bib/longstrings,main}
}


\clearpage
\appendix

\input{sections/A_IntroSupp}
\section{Dataset Details}
\label{sup_sec:dataset}
We start with the essential step of creating soft labels for each shot and dialog of the episode, presented in \cref{sup_sec:matching}.
We describe how recaps are used to generate binary labels using our novel shot-matching algorithm.
Next, \cref{sup_sec:smoothing} describes how label smoothing is performed.
This is an essential step towards capturing the local sub-story in a better way.
Further, \cref{sup_sec:splits} comments on the different data split creation strategies used to evaluate baselines and our model's generalizability.
Finally, to conclude \cref{sup_sec:lab_rel} presents detailed episode-level label reliability scores.
\input{sections/A_Matching}

\input{sections/B_Smoothing}
\input{sections/G_Misc}

\section{Experiments and Results}
\label{sup_sec:expt_res}
In this section
we expand our implementation details (\cref{sup_sec:impl_details}) and present additional details of the backbones used for feature extraction (\cref{sup_sec:feat_extraction}).
Extensive ablation studies with regard to feature combinations and model hyperparameters are presented in~\cref{sup_sec:ablations}.

In~\cref{sup_sec:adapt}, we present details of the modifications that need to be made to adapt SoTA baselines such as MSVA~\cite{msva}, PGL-SUM~\cite{pglsum}, PreSumm~\cite{presumm}, and A2Summ~\cite{he2023a2summ} for comparison with \modelname{}.
\cref{sup_sec:bench_data} details how we adapted our model for SumMe and TVSum, along with comprehensive hyperparameter particulars.
To conclude, \cref{sup_sec:qa} shows extended qualitative analysis, as in the main paper, on three other episodes (S06E20, S07E22, and S05E21).

\input{sections/C0_ImplementationDetails}
\input{sections/C_FeatExtract}
\input{sections/D_Ablations}

\input{sections/E_SoTA_Adapt}
\input{sections/F_Benchmark}
\input{sections/H_QA}

\input{sections/I_Limitations}

\end{document}

%% file: preamble.tex
\usepackage{graphicx}
\usepackage{amsmath}
\usepackage{amssymb}
\usepackage{bbm}
\usepackage{booktabs}
\usepackage{caption}
\usepackage{mathrsfs}
\usepackage{mathtools}
\usepackage{rotating}
\usepackage{multirow}
\usepackage{lipsum}
\usepackage[dvipsnames]{xcolor}
\usepackage{enumitem}
\usepackage{inconsolata}
\usepackage{pifont}
\usepackage{soul}
\usepackage{siunitx}
\usepackage{subcaption}
\usepackage{float}
\usepackage[accsupp]{axessibility}  
\usepackage{xcolor}
\usepackage{balance}
\usepackage{makecell}

\newcommand{\modelname}{TaleSumm}
\newcommand{\dataname}{PlotSnap}

\renewcommand{\paragraph}[1]{\vspace{1mm}\noindent\textbf{#1}}

\newcommand{\intracvt}{IntraCVT}
\newcommand{\xseason}{X-Season}
\newcommand{\xseries}{X-Series}
\newcommand{\tf}{\emph{24}}
\newcommand{\pb}{\emph{Prison Break}}

\newcommand{\STRM}{\mathsf{ST}}
\newcommand{\ETRM}{\mathsf{ET}}
\newcommand{\soft}{\mathsf{softmax}}
\newcommand{\bbf}{\mathbf{f}}
\newcommand{\bu}{\mathbf{u}}

\newcommand{\by}{\mathbf{y}}
\newcommand{\bw}{\mathbf{w}}

\newcommand{\bs}{\mathbf{s}}
\newcommand{\bo}{\mathbf{o}}
\newcommand{\bq}{\mathbf{q}}

\newcommand{\W}{\mathbf{W}}
\newcommand{\bW}{\mathbf{W}}
\newcommand{\F}{\mathbf{F}}

\newcommand{\bE}{\mathbf{E}}
\newcommand{\bA}{\mathbf{A}}

\newcommand{\CLS}{\mathsf{CLS}}
\newcommand{\BCE}{\mathsf{BCE}}

\newcommand{\bbR}{\mathbb{R}}

\newcommand{\MN}{\mathcal{N}}
\newcommand{\MU}{\mathcal{U}}

\newcommand{\MS}{\mathcal{S}}

\newcommand{\ME}{\mathcal{E}}

\newcommand{\MR}{\mathcal{R}}
\newcommand{\MP}{\mathcal{P}}

\newcommand{\ML}{\mathcal{L}}
\newcommand{\MG}{\mathcal{G}}

\newcommand{\dingcheck}{\ding{51}}
\definecolor{darkred}{rgb}{0.5, 0.0, 0.13}
\definecolor{royalblue}{HTML}{002855}
\definecolor{yorange}{HTML}{FFA445}
\definecolor{g1}{HTML}{2A7208}
\definecolor{g2}{HTML}{4A9C13}
\definecolor{g3}{HTML}{82C720}
\definecolor{purpletext}{HTML}{8856A7}


%% file: sections/00_abstract.tex
\label{sec:abs}
\begin{abstract}
\vspace{-2mm}
We introduce \emph{multimodal story summarization} by leveraging TV episode recaps -- short video sequences interweaving key story moments from previous episodes to bring viewers up to speed.
We propose \emph{\dataname{}}, a dataset featuring two crime thriller TV shows with rich recaps and long episodes of 40 minutes.
Story summarization labels are unlocked by matching recap shots to corresponding sub-stories in the episode.
We propose a hierarchical model \emph{\modelname{}} that
processes entire episodes by creating compact shot and dialog representations, and
predicts importance scores for each video shot and dialog utterance by enabling interactions between local story groups.
Unlike traditional summarization, our method extracts multiple plot points from long videos.
We present a thorough evaluation on story summarization, including promising cross-series generalization.
\modelname{} also shows good results on classic video summarization benchmarks.

\end{abstract}

%% file: sections/01_introduction.tex
\vspace{-3mm}
\section{Introduction}
\label{sec:intro}

Imagine settling in to catch the latest episode of our favorite TV series.
We hit play and the familiar ``\emph{Previously on ...}'',
the recap, a smartly edited segment swiftly brings us up to speed, reminding us of key moments from past episodes.

A TV show \textbf{recap} is a concise, under-two-minute sequence of crucial plot points from previous episodes.
To satisfy the time constraint, the recap is constructed by editing shots from previous episode with sharp and rapid cuts
and selecting/modifying dialog utterances to ensure relevance to the sub-story.
A good recap sets the stage for the main part of the episode by weaving visual and dialog cues to \emph{spark the viewers' memory}.
Thus, a recap is a great way to identify sub-stories important to the overall story arc.

We use recaps to create \textbf{story summaries} by identifying and expanding the sub-stories from the episode 
(\cref{fig:teaser}).

We introduce an innovative shot-matching algorithm (\cref{sec:dataset}) that associates shots from the recap to their corresponding shots in the episode.
Different from a recap, a story summary consists of entire scenes or sub-stories that are essential to the narrative.
Thus, a first-time viewer may watch 
story summaries of each episode serially and understand the main narrative,
while watching recaps serially does not help as they are only meant as memory triggers and assume that the viewer has seen the episode before.

\input{sections/tables/data_compare}

We propose a novel task of \textbf{creating multimodal story summaries} for TV episodes.
We introduce \emph{\dataname{}}, a new dataset for story summarization consisting of two popular crime thrillers:
(i)~\tf{}~\cite{24} features Jack Bauer, an agent at the counter-terrorism unit who relentlessly tackles seemingly impossible missions; and
(ii)~\pb{}~\cite{pb} features Michael Scofield who plans and executes daring escapes from prisons.
We choose action thrillers as they are often more challenging than romantic and situational comedies with multiple suspenseful story-lines, rapid action sequences, and complex visual scenes.
With excellent recaps in both shows, we can extract important narrative subplots from the recap to create story summaries (see ~\cref{sec:dataset}).

Our task of story summarization is an instance of multimodal long-video understanding where an entire episode (typically 40 minutes) needs to be processed.
We formulate story summarization as an extractive multimodal summarization task with multimodal outputs (video-text to video-text, VT2VT).
Specifically, we build models that predict the importance of each video shot and dialog utterance (story elements) in an episode.
Selecting multiple major and connected sub-stories is different and challenging from most summarization works that promote visual diversity~\cite{dpp}.

We also propose a new hierarchical Transformer model, \modelname{}, to perform story summarization.
Different from typical summarization approaches~\cite{zhu-etal-2018-msmo-1,fu-etal-2021-msmo-2,papalampidi2022hier3d} that use multimodal inputs to either generate a video (select frames) \emph{or} a text summary, our model predicts scores for both modalities.
Recent multimodal approaches, A2Summ~\cite{he2023a2summ} and VideoXum~\cite{lin2023videoxum}, also generates both outputs; but we differ significantly in video type (stories \vs~creative videos), the duration of the input video, and the model architecture.
The first level of our model encodes shot and utterance representations.
At the second level, we foster interaction between shots and utterances within local story groups based on a temporal neighborhood, reducing the impact of distant and potentially noisy elements.
A dedicated group token enables message-passing across story groups.

In summary, our contributions are:
(i)~We propose story summarization that requires identifying and extracting multiple plot points from narrative content.
This is a challenging multimodal long-video understanding task. 
(ii)~We pioneer the use of TV show recaps for video understanding and show their application in story summarization.
We introduce \dataname{}, a new dataset featuring 2 crime thriller TV series with rich recaps.
(iii)~We propose a novel hierarchical model that features shot and dialog level encoders that feed into an episode-level Transformer.
The model operates on the full episode while being lightweight enough to train on consumer GPUs.
(iv)~We present an extensive evaluation: ablation studies validate our design choices, \modelname{} obtains SoTA on \dataname{} and performs well on video summarization benchmarks.
We show generalization across seasons and even across TV shows, and evaluate consistency of labels obtained from multiple diverse sources.

%% file: sections/tables/data_compare.tex
\begin{table*}[t]
\centering
\footnotesize
\tabcolsep=0.06cm
\begin{tabular}{l c c c l l}
\toprule
Dataset                                   & Modalities & \#    & Length    & \multicolumn{1}{c}{Content}                   & \multicolumn{1}{c}{Summary Annotations} \\ \midrule
SumMe~\cite{summe}                        & V2V        & 25    & 1-6 min   & Holidays, events, sports                      & Multiple set of key fragments           \\
TVSum~\cite{tvsum}                        & V2V        & 50    & 1-11 min  & News, how-to, user-generated, documentary     & Multiple fragment-level scores          \\
OVP~\cite{avila2011vsumm}                 & V2V        & 50    & 1-4 min   & Documentary, educational, historical, lecture & Multiple set of key-frames              \\ \midrule
CNN-DailyMail~\cite{cnn_daily} & T2T      & 311672 & 766 words   & News articles and highlight stories           & Human-generated internet summaries      \\
XSum~\cite{xsum} & T2T & \multicolumn{1}{l}{226711} & 431 words & BBC News articles & Single-sentence summary by author \\
TRIPOD~\cite{tripod}                      & T2T        & 99    & ~22072 words   & Movies (action, romance, comedy, drama)       & Synopses level annotations              \\
SummScreen~\cite{summscreen} & T2T & 26851 & 7013 words & TV screenplays (wide scope, 21 genres) & Human-written internet summaries \\ \midrule
How2~\cite{how2}                          & VT2T       & 79114 & 2-3 min   & Instructional videos                          & Youtube descriptions (and translations) \\
SummScreen3D~\cite{papalampidi2022hier3d} & VT2T       & 4575  & ~5721 words   & TV Shows (soap operas)                        & Human written internet summaries        \\
BLiSS~\cite{he2023a2summ} & VT2VT       & 13303  & ~10.1 min/49 words   & Livestream Videos          & Human text-summaries; Thumbnail animation \\
\textbf{\dataname{} (Ours)}                   & VT2VT      & 215   & 40-45 min & TV Shows (crime thriller)                     & Matching recap shots followed by smoothing      \\
\bottomrule
\end{tabular}
\vspace{-2mm}
\caption{
Overview of video/text/multimodal summarization datasets.
\# indicates the size of the dataset (no. of instances).
The modalities column includes:
V2V: Video-to-video,
T2T: Text-to-text,
VT2T: Video-text to text, and
VT2VT: Video-text to video-text summarization.
Closest to our domain of story summarization are SummScreen and SummScreen3D, however, they produce text summaries.
}
\vspace{-4mm}
\label{tab:data_compare}
\end{table*}

%% file: sections/02_related_works.tex
\section{Related Work}
\label{sec:related_works}

Video summarization predates Deep Learning (DL).
Past methods focused on
generating 
keyframes~\cite{keyframes1,keyframes2,keyframes3,keyframes4},
skims~\cite{skims2,interestingness1},
video storyboards~\cite{storyboards},
time-lapses~\cite{time-lapses},
montages~\cite{montages}, or
video synopses~\cite{synopses}.
However, given the effectiveness of DL methods (\eg~\cite{conv1,gygli,conv3,conv4}) over traditional optimization-based approaches, we will primarily discuss learning-based approaches in the following.

\paragraph{Summarization modalities.}
We classify approaches based on input and output modalities.
\textbf{(i)}~Video to frames/video (V2V) approaches
model temporal relations~\cite{dsnet,rrstg,reseq2seq},
preserve diversity~\cite{temporal1,dpp},
or generate images/videos~\cite{supgan1,supgan2,gan9,trecvid}.
On the other, \textbf{(ii)}~text to text (T2T) methods are either
\emph{extractive}~\cite{presumm,matchsum,hahsum} picking important sentences from a document, or
\emph{abstractive}~\cite{bart,t5,pegasus} summarizing the overall meaning by generating new text~\cite{nallapati2016abstractiveseq2seq}.
Relevant to our work, story screenplay summaries~\cite{papalampidi2020screenplaysummary, summscreen} or turning point identification~\cite{tripod} can be seen as T2T summarization.

Multimodal approaches typically benefit from additional modalities to enhance model performance.
\textbf{(iii)}~Video-text to text (VT2T) is popular for screenplays~\cite{papalampidi2022hier3d,tripodplus}, particularly in generating video captions~\cite{tacos, how2, trecvid}.
\textbf{(iv)}~Video-text to video (VT2V) covers the field of \emph{query-guided summarization}~\cite{query1,query2,query3}.
Finally, the last option is \textbf{(v)}~video-text to video-text (VT2VT) summarization.
Our work lies here and is different from A2Summ~\cite{he2023a2summ} and VideoXum~\cite{lin2023videoxum}, as we operate on long videos edited to convey complex stories.
Different from trailer generation~\cite{papalampidi2021trailer} that avoids spoilers, we wish to identify all key story events.

\paragraph{Summarization datasets.}
We compare popular summarization datasets based on above modalities in Table~\ref{tab:data_compare}.
Video-only datasets, TVSum~\cite{tvsum} and SumME~\cite{summe}, consist of short duration videos unlike ours.
Other video datasets work with
first-person videos~\cite{fvpsum},
are used for title generation~\cite{vtw}, and even
feature e-sports audience reactions~\cite{lol}.
For a nice overview of text-only (T2T) and text-primary (VT2T) datasets, we refer the reader to~\cite{papalampidi2022hier3d}.
Briefly, text datasets include
news articles (CNN-DailyMail~\cite{cnn_daily}, XSum~\cite{xsum}),
human dialog (Samsum~\cite{samsum}), and
TV/movie screenplays (SummScreen~\cite{summscreen}).
While similar in spirit to screenplays used for storytelling, \dataname{} is different as it features TV episodes with long videos and dialogs (without speaker labels or scene descriptions), a significant challenge in long-form video understanding.

\paragraph{Story summarization}
retrieves multiple sub-stories contained within the story-arc of an episode.
To our best knowledge, we are unaware of works on video-text story-summary generation.
There are attempts to understand stories in movies/TV shows through various dimensions:
person identification~\cite{tapaswiNamingChars, Nagrani2018LearnablePCPIN, Nagrani2018FromBC2Sherlock, HanAutoAD2},
question-answering~\cite{tapaswiMovieQA, lei-etal-2018-tvqa, lei-etal-2020-tvqa},
captioning~\cite{Rohrbach2015ADF_moviedesc, lsmdc, lsmdc_fill_in_2020, mvad},
situation understanding~\cite{moviegraphs, vidsitu, khan2022grounded, XiaoVidsitu2},
text alignment~\cite{tapaswi2015AlignPlot, tapaswiStoryBased, Zhu2015AligningBA, Sankar2009SubtitlefreeMT}, or
scene detection~\cite{s4scenedetection, s4clip, movie2scene, hao2021recapdet}.
Recently, SummScreen3D~\cite{papalampidi2022hier3d} extends SummScreen~\cite{summscreen} with visual inputs, but the output summary is still textual.
On the other hand, our goal is multimodal story-summary generation by predicting both - important video shots \emph{and} dialogs.

%% file: sections/03_dataset.tex
\section{\dataname{} Dataset}
\label{sec:dataset}

We introduce the \dataname{} dataset consisting of long-form multimodal TV episodes with a well-structured underlying plot spanning multiple seasons and episodes.
We consider two American crime thriller TV shows with rich storylines:
\tf{}~\cite{24} and \pb{}~(PB)~\cite{pb}.
Unlike sitcoms, crime thrillers are recognized for their methodically crafted captivating plot lines.
Notably, both \tf{} and \pb{} have good recaps, and are famous for using the catchphrase ``\emph{Previously on ...}'' at the start of the recap.

We present some statistics of \dataname{} in Table~\ref{tab:data_stat}.
With a total of 205 episodes, the large number of shots and dialogs present in each episode pose interesting challenges for summarization.
The first section of the table presents overall size and duration,
second shows statistics for shots and dialog utterances, and the third for recaps.
We note that recap shots are much shorter (1.9s \vs~3.2s for \tf) allowing the editors to pack more story content in the same duration.

\input{sections/tables/data_stats}

\paragraph{Our key idea} is to use professionally edited recaps, shown at the beginning of a new episode, as labels for story summarization.
Let $\ME_n$ be the $n^\text{th}$ episode in a TV series.
$\MR_{n+1}$ is the recap shown just before the episode $\ME_{n+1}$ begins and may contain content from all past episodes $\{\ME_n,\ldots,\ME_1\}$.
Thus, we classify the visual content appearing in the recap into three sources along with their average proportions (for \tf):
(i)~shots that are picked (and usually trimmed) from $\ME_n$ (88\%),
(ii)~shots that are picked from $\ME_{n-1}$ or earlier (5\%), and
(iii)~new shots that did not appear in any previous episode (7\%).
As most shots (88\%) of a recap are from the preceding episode, recaps serve as good summary labels.
The remaining 12\% recap content (from earlier episodes or unseen shots) is ignored.
We also remove the last episode of each season due to the absence of a recap.

\paragraph{Recap inspired labels.}
We present how recap shots and dialogs can be used to create labels for story summarization.
First, we manually extract the recap ($\MR_{n+1}$) from $\ME_{n+1}$ instead of employing automatic detection methods~\cite{hao2021recapdet} to avoid introducing additional label noise.
Second, to localize trimmed recap shots in past episodes ($\ME_{1},\ldots,\ME_{n}$), we propose a shot-matching algorithm that conducts pairwise comparisons of frame-level embeddings, making selections based on a threshold determined by similarity score and frequency.
Due to \textit{shot thread} patterns~\cite{tapaswi2014storygraphs}, one recap shot may match multiple shots in the episode.
This is desirable as we want to highlight larger sub-stories as part of the summary.
In fact, selecting only one shot in a thread adversely affects the model due to conflicting signals as shots with very similar appearance are assigned opposite labels.

We think of recap matched shots as temporal point annotations~\cite{label_sup}.
We identify the set of matching shots in the episode,
create a binary label vector,
and smooth this vector using a triangular filter.
We will refer to these smoothed labels as ground-truth (GT) for story summarization.
Extending the supervision helps the model identify meaningful, contiguous sub-stories rather than focusing solely on specific shots highlighted in the recap.
For example, it is unlikely that shot $s_i$ is important to the story while $s_{i\pm1}$ is entirely irrelevant (except at scene boundaries).
Thus, smoothing is essential to clarify the distinction between positive (essential) and negative (unimportant) shots.
Please refer to \cref{sup_sec:matching,sup_sec:smoothing} for details on the label creation process.

A similar approach can be adopted for dialog utterances.
We are able to match 88\% of recap utterances to dialog within the smoothed video labels.
The rest do not appear in episode $\ME_n$ or are picked from extra recorded footage.
For simplicity, we inherit labels for the dialogs based on the smoothed label for the temporally co-occurring shot.

%% file: sections/tables/data_stats.tex
\begin{table}[t]
\centering
\small
\tabcolsep=0.12cm
\begin{tabular}{l c c}
\toprule
\multicolumn{1}{c}{TV Series} & 24 & Prison Break \\
\midrule
\# of Seasons & 8 & 2 \\
\# of Episodes & 172 & 33 \\
Dataset duration (hours) & 125.9 &  24.0 \\
Avg episode duration (s) & 2635 $\pm$ 72 & 2615 $\pm$ 39 \\
\midrule
Avg \# of shots per episode & 825 $\pm$ 101 & 999 $\pm$ 117 \\
Avg duration of shots (s) & 3.2 $\pm$ 2.5 & 2.6 $\pm$ 2.3 \\
Avg \# of utterances per episode & 564 $\pm$ 54 & 529 $\pm$ 59 \\
Avg \# of words/tokens in utterance & 7.9 $\pm$ 5.4 & 7.4 $\pm$ 5.8 \\
\midrule
Avg recap duration (s) & 104 $\pm$ 28 & 62 $\pm$ 20 \\
Avg \# of shots in recap & 55 $\pm$ 12 & 43 $\pm$ 9 \\
Avg \# of utterances in recap & 33 $\pm$ 6 & 22 $\pm$ 5 \\
\bottomrule
\end{tabular}
\vspace{-2mm}
\caption{Mean ($\pm$ stddev) featuring properties of video shots, dialog utterances, and the recap in our dataset \dataname{}.}
\vspace{-5mm}
\label{tab:data_stat}
\end{table}

%% file: sections/04_method.tex
\section{Method: \modelname{}}
\label{sec:method}

\begin{figure*}
\centering
\includegraphics[width=\linewidth]{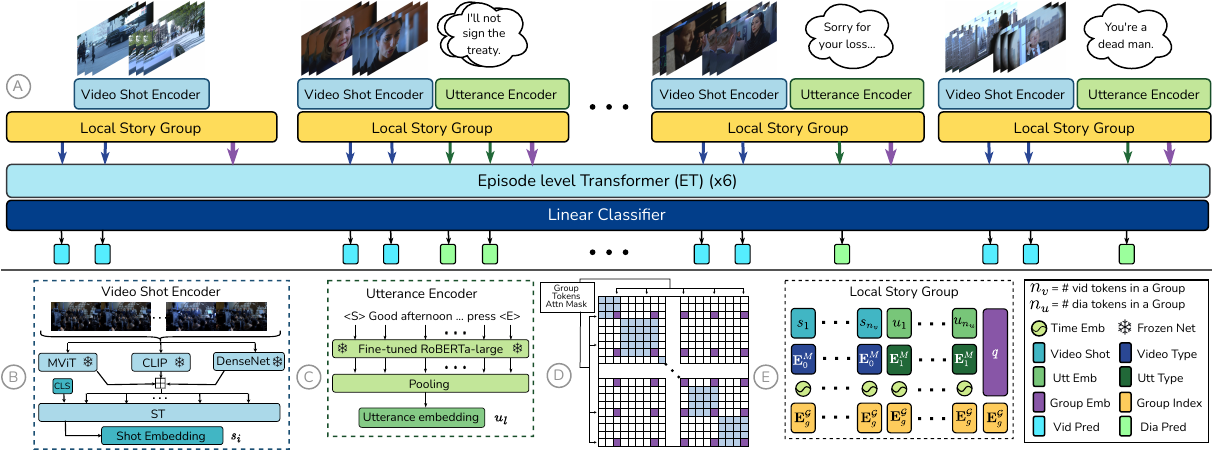}
\vspace{-4mm}
\caption{
\textbf{(A)~\modelname{}} ingests all video shots and dialogs of the episode and encodes them using (B) and (C).
Based on temporal order, we combine tokens into local story groups (\emph{illustration} shows small groups of 2 shots and 0-2 utterances).
To each group, we append a {\color{purpletext} group token} and add multiple embeddings, before feeding them to the the episode-level Transformer ($\ETRM$).
For each shot or dialog token, a linear classifier predicts its importance.
\textbf{(B)~Video shot encoder.}
For each frame, representations from multiple backbones are fused using attention ($\boxplus$).
We feed these to a shot Transformer encoder $\STRM$, and tap a shot-level representation from the $\CLS$ token.
\textbf{(C)~Utterance encoder} uses a fine-tuned language model and avg-pooling across all words of the utterance.
\textbf{(D)~Self-attention mask} illustrates the block-diagonal self-attention structure across the episode. Group tokens across the episode (purple squares) communicate with each other. 
\textbf{(E)~Multiple embeddings} are added to the tokens to capture modality type, time, and membership to a local story group.}
\vspace{-2mm}
\label{fig:method}
\end{figure*}

We introduce \modelname{}, a two-level hierarchical model that identifies important sub-stories in a TV episode's narrative (illustrated in Fig.~\ref{fig:method}).
At the first level, our approach exploits frame-level (word-level) interactions to extract shot (dialog) representations (Sec.~\ref{subsec:level1}, Fig.~\ref{fig:method}(B, C)).
At the second level, we capture cross-modal interactions across the entire episode through a Transformer encoder (Sec.~\ref{subsec:level2}, Fig.~\ref{fig:method}(A, D)).
Before diving into the architecture, we formalize story summarization and introduce notation.

\subsection{Problem Statement}
\label{subsec:prob_state}

Our aim is to extract a multimodal story summary (video and text) from a given episode, typically lasting around 40 minutes, and encompassing multiple key events.

\paragraph{Notation.}
An episode $\ME = (\MS, \MU)$ consists of
a set of $N$ video shots $\MS = \{s_i\}_{i=1}^{N}$ and
a set of dialog utterances $\MU = \{u_l\}_{l=1}^{M}$.
A \emph{shot} serves as a basic unit of video processing and comprises temporally contiguous frames taken from the same camera, while a \emph{dialog utterance} typically refers to a sentence uttered by an individual as part of a larger conversation.
We denote each shot as $s_i = \{f_{ij}\}_{j=1}^{T_i}$, where $f_{ij}$ are sub-sampled frames, and
each utterance as $u_l = \{w_{lp}\}_{p=1}^{T_l}$ with multiple word tokens $w_{lp}$.

\paragraph{Summarization as importance scoring.}
While humans may naturally select start and end temporal boundaries to indicate important sub-stories, for ease of computation, we discretize time and associate an importance score with each video shot or dialog utterance.
Thus, given an episode $\ME = (\MS, \MU)$, we formulate story summarization as a binary classification task applied to each element (shot or dialog).
The ground-truth labels can be denoted as
$\by^{\MS} = \{y_i^{\MS}\}_{i=1}^{N}$ and
$\by^{\MU} = \{y_l^{\MU}\}_{l=1}^{M}$, where each
$y_i^{\MS}, y_l^{\MU} \in [0, 1]$, 
signaling their importance to the story summary.

\subsection{Level 1: Shot and Dialog Representations}
\label{subsec:level1}

In narrative video production, \emph{shots} play an important role in advancing the storyline and contextualizing neighboring content.
We obtain shot-level representations from granular frame-level features to determine how well the shot can contribute to understanding the storyline.

\paragraph{Feature extraction.}
To capture various aspects of the shot, we use \emph{three} pretrained backbones that capture visual diversity through people, their actions, objects, places, and scenes: $\phi_{\MS}^{k}(\cdot), k = \{1, 2, 3\}$.
We extract relevant visual information from frame(s) of a given shot, $s_i$ as follows:
\begin{equation}
\bbf_{ij}^k = \phi_\MS^k \left( \{f_{ij}\} \right) \, , \quad 
\bbf_{ij}^k \in \bbR^{D_\MS^k}.
\end{equation}
Note that the backbone may encode a single frame $f_{ij}$ or a short sequence around $f_{ij}$.

For dialog utterances, we adopt a fine-tuned
language model $\phi_{\MU}^\text{FT}$, to compute contextual word-level features:
\begin{equation}
\bw_{lp} = \phi_{\MU}^\text{FT} \left( \{ w_{lp} \} \right) \, , \quad \bw_{lp} \in \bbR^{D_{\MU}}  \, .
\end{equation}

\paragraph{Shot $\CLS$ pooling.}
To compute an aggregated shot representation, we combine frame-level signals into a compact
representation.
An attention-based aggregation ($\boxplus$) (inspired by~\cite{lstm}), 
effectively weighs the most pertinent information (\eg~action in a motion-heavy shot or scenery in an establishing shot).
First, the frame features from different backbones are projected to the same space (using $\W_{\MS}^{k}{\in}\ \bbR^{D\times D_{\MS}^{k}}$) and then concatenated to form $\hat{\bbf}_{ij}^{1:3} \in \bbR^{3D}$~(Eq.~\ref{eq:SS}).
A linear layer $\W_{\MP}\in \bbR^{3\times 3D}$ followed by $\tanh$ and $\soft$ computes scalar importance scores that are used for weighted fusion:
\begin{align}
\label{eq:SS}
\hat{\bbf}_{ij}^{1:3} &= [W_\MS^{1}\bbf_{ij}^{1}, W_\MS^{2}\bbf_{ij}^{2}, W_\MS^{3}\bbf_{ij}^{3}] \, , \\
\boldsymbol{\alpha}_{ij}^{1:3} &= \soft(\tanh(W_\MP \hat{\bbf}_{ij}^{1:3})) \, , \\
\F_{ij} &= \alpha_{ij}^{1} \hat{\bbf}_{ij}^1 + \alpha_{ij}^{2} \hat{\bbf}_{ij}^2 + \alpha_{ij}^{3} \hat{\bbf}_{ij}^3 \, .
\end{align}
We omit bias for brevity.
We add relative frame position to $\F_{ij}$ through a \emph{time-embedding} vector, $\bE_{j}^{\MS}$, similar to Fourier position encoding~\cite{attention}.

A shot transformer~\cite{attention} $\STRM$ is used to encode the frame-level feature sequence $\{\F_{ij}\}_{j=1}^{T_i}$.
We tap the output from the $\CLS$ token appended at the beginning of the sequence (\eg~similar to BERT~\cite{bert}) as the final shot representation:
\begin{equation}
\bs_i = \STRM( \{ \F_{ij} + \bE_{j}^{\MS} \}_{j=1}^{T_i} ) \, , \quad \bs_i \in \bbR^D \, .
\end{equation}

\paragraph{Dialog utterance representation.}
First, we project the tokens $\bw_{lp}$ to $\bbR^{D}$ using a linear layer $\W_{\MU}\in \bbR^{D\times D_{\MU}}$.
As the tokens are already contextualized by $\phi_\MU^\text{FT}$, a simple mean-pool across the $p$ tokens is found to work well:
\begin{equation}
\bu_l = \text{mean}_p ( \{\W_{\MU} \bw_{lp} \}_{p=1}^{T_l} ) \, , \quad \bu_l \in \bbR^D \, .
\end{equation}

\subsection{Level 2: Episode-level Interactions}
\label{subsec:level2}
We propose an episode-level Transformer encoder, $\ETRM$, that models interactions across shots and dialog of the entire episode.
Predicting the importance of an element (shot or dialog) requires context in a neighborhood; \eg~shot in a scene, dialog utterance in a conversation.

\paragraph{Additional embeddings.}
We arrange shot and dialog tokens based on their order in the episode (see Fig.~\ref{fig:method}(A)).
Learnable \emph{type embeddings} help the model distinguish between shot and dialog modalities ($\bE^{M}\in \bbR^{2\times D}$).
We encode the real time (in seconds) of appearance of each element (shot or dialog) using a binning strategy.
Given an episode of $T$ seconds, we initialize Fourier position encodings $\bE^T \in \bbR^{\lceil T/\tau \rceil \times D}$ where $\tau$ is the bin-size.
Based on the mid-timestamp of each element $t$, we add $\bE_t^T$ to the representation, the $\lfloor t/\tau \rfloor^\text{th}$ row in the position encoding matrix.
Such time embeddings allow our model to:
(i)~implicitly encode shot duration; and
(ii)~relate co-occurring dialogs with video shots without the need for complex attention maps.

\paragraph{Local story groups.}
The total number of video shots and dialog that make up the sequence length for $\ETRM$ is $S{=}N{+}M$ ($\sim$1500).
Self-attention over so many tokens is not only computationally demanding, but also difficult to train due to unrelated temporally distant tokens that happen to look similar.
We adopt a block-diagonal attention mask to constrain the tokens to attend to local story regions:
\begin{equation}
\bA_{S \times S} = \text{diag}( \mathbbm{1}_{n_1\times n_1}, \ldots, \mathbbm{1}_{n_g\times n_g}, \ldots, \mathbbm{1}_{n_G \times n_G} ) \, ,
\end{equation}
where $\mathbbm{1}_{n_g\times n_g}$ denotes an all one matrix,
$n_g$ is the \# of tokens in the $g^\text{th}$ local block,
$\sum_{g=1}^{G} n_g = S$, and
$\text{diag}(\ldots)$ constructs a block diagonal matrix.
We add new learnable group index embeddings $\bE^{\MG} \in \bbR^{G\times D}$ to our tokens to inform our model about their group membership.

\paragraph{Group tokens.}
While capturing interactions across all tokens may lead to poor performance, self-attention only within the local story groups prohibits the model from capturing long-range story dependencies.
To enable story group interactions, we propose to add a set of \emph{group tokens} to the input, extending the sequence length to $\hat{S}{=}S{+}G$.
The group tokens $\bq_g$ represent an additional layer of hierarchy within the episode model as they summarize the story content inside a group and also communicate across groups, providing a way to understand the continuity of the story.
Fig.~\ref{fig:method}(E) shows how group tokens are inserted at the end of each local story group's shot and dialog tokens.

To facilitate cross-group communication, we make two modifications to the self-attention mask:
(i)~The size of each local group $n_g$ is extended by 1 to incorporate the group token $\bq_g$ within the block matrix.
We also update $\bA$ to reflect this and is of size $\hat{S} \times \hat{S}$.
(ii)~We compute a binary index $\bo \in \{0, 1\}^{\hat{S}}$ to represent the locations at which a group token appears in the sequence.
The new self-attention mask $\hat{\bA} = \bA + \bo \bo^T$ allows group-tokens to communicate.
Fig.~\ref{fig:method}(D) illustrates the attention mask; light blue squares correspond to attention within a group, and sparse purple squares visualize attention across the group tokens.

\paragraph{Importance prediction.}
We present how shot or dialog scores can be estimated.
First, the input tokens to $\ETRM$ are:
\begin{eqnarray}
\hat{\bs}_i &=& \bs_i + \bE^M_0 + \bE^T_{t_i} + \bE^\MG_{g_i} \, , \\
\hat{\bu}_l &=& \bu_l + \bE^M_1 + \bE^T_{t_l} + \bE^\MG_{g_l} \, , \\
\bq_g &=& \bq + \bE^\MG_g \, .
\end{eqnarray}
where $t_i, t_l$ and $g_i, g_l$ correspond to the mid-timestamp and group membership of shot $s_i$ and dialog $u_l$ respectively.
$\bq$ denotes the learnable shared group type embedding.

We feed the updated shot, dialog, and group token representations to $\ETRM$ post LayerNorm~\cite{layernorm}, a $H_{\mathsf{E}}$ layer Transformer encoder with a curated self-attention mask $\hat{\bA}$:
\begin{equation}
[\ldots, \tilde{\bs}_i, \tilde{\bu}_l, \tilde{\bq}_g, \ldots] =
\ETRM(
[\ldots, \hat{\bs}_i,   \hat{\bu}_l,   \bq_g, \ldots]; \hat{\bA}
) \, ,
\end{equation}
with all tokens, \ie~$\{i\}_1^N$, $\{l\}_1^M$, and $\{g\}_1^G$.

After $\ETRM$, we compute shot and dialog importance scores using a shared linear classifier $\W^{C} \in \bbR^{1\times D}$ followed by sigmoid function $\sigma(\cdot)$:
\begin{equation}
\label{eq:imp_scores}
\hat{y}^\MS_i = \sigma(\W^{C} \tilde{\bs}_i)
\, \text{ and } \,
\hat{y}^\MU_l = \sigma(\W^{C} \tilde{\bu}_l)
\, \forall i,l \, .
\end{equation}

\subsection{Training and Inference}
\label{subsec:train}

\paragraph{Training.}
\modelname{} is trained in an end-to-end fashion with \emph{BinaryCrossEntropy} ($\BCE$) loss. We provide positive weights, $w$ (ratio of negatives to positives) to account for class imbalance.
Modality specific losses are added:
\begin{equation}
\ML = \BCE\left(\hat{\by}^{\MS}, \by^{\MS}; w^\MS\right) + \BCE\left(\hat{\by}^{\MU}, \by^{\MU}; w^\MU \right) \, .
\end{equation}

\paragraph{Inference.}
At test time, we follow the procedure outlined in Sec.~\ref{subsec:level2} and generate importance scores for each video shot and dialog utterance (Eq. \ref{eq:imp_scores}).

\paragraph{Model ablations.}
As we will see empirically, our model is versatile and well-suited for adding/removing modalities or additional representations by adjusting the sequence length of the Transformer (number of tokens).
It can also be modified to act as an unimodal model that applies only to video or dialog utterances by disregarding other modalities.

%% file: sections/05_experiments.tex
\section{Experiments and Analysis}
\label{sec:expt}

We first discuss the experimental setup.

\paragraph{Data splits.}
We adopt 3 settings.
(i)~\intracvt: On \tf, most experiments (\cref{tab:vid_dia_abl,tab:mm_ablations,tab:sota_abl}) follow an \underline{intra}-season 5-fold \underline{c}ross-\underline{v}alidation-\underline{t}est strategy.
(ii)~\xseason: On \tf, we assess cross-season generalization using a 7-fold cross-validation-test (\cref{tab:split_table}).
(iii)~\xseries: shows transfer results from \tf{} to \textit{PB} (\cref{tab:split_table}).
More details can be found in \Cref{sup_sec:splits}.

\paragraph{Evaluation metric.}
We adopt Average Precision (AP, area under PR curve) as the metric to compare predicted importance scores of shots or dialogs against ground-truth.
\subsection{Implementation details}
\label{subsec:impl_details}

We present some high-level details here.

\paragraph{Feature backbones.}
We adopt three visual backbones:
DenseNet169~\cite{densenet} for object understanding;
MViT~\cite{mvit} for action information; and
OpenAI CLIP~\cite{clip} for semantics.

To encode dialog, we adapt $\text{RoBERTa-large}$~\cite{roberta} for extractive summarization using parameter-efficient fine-tuning~\cite{adaptor, papalampidi2022hier3d} on the text from our dataset.
The backbone is frozen when training \modelname{} for story summarization.

Additional backbone details are in \Cref{sup_sec:impl_details,sup_sec:feat_extraction}.

\paragraph{Architecture.}
We find $H_{\mathsf{S}}{=}1$, $H_{\mathsf{E}}{=}6$, and $n_g{=}20$ to work best.
Both $\STRM$ and $\ETRM$ have the same configuration: 8 attention heads and $D{=}128$.
We tried several architecture configurations whose details can be found at \Cref{sup_sec:impl_details}.

\paragraph{Training details.}
We \emph{randomly} sample up to 25 frames per shot during training as a form of data augmentation and use \emph{uniform} sampling during inference.
Our model is implemented in PyTorch~\cite{pytorch}, has 1.94 M parameters,
and is trained on 4 RTX-2080 Ti GPUs for a batch size of 4 (\ie~4 entire episodes).
The optimizer, learning rate, dropout, and other hyperparameters are tuned for best performance on the validation set and indicated in \Cref{sup_sec:impl_details}.

\subsection{Experiments on \tf}
\label{subsec:abl}

\input{sections/tables/vid_dia_ablations}

\paragraph{Architecture ablations.}
Results in \cref{tab:vid_dia_abl} are across two dimensions:
(i)~columns span the shot or utterance level and
(ii)~rows span the episode level.
All model variants outperform a baseline that predicts a random score between $[0, 1]$: AP 34.2 (video) and 30.4 (dialog),  over 1000 trials.

Across rows, we observe that the MLP performs worse than the other two variants by almost 10\% AP score
because assuming independence between story elements is bad.
Our proposed approach with local story groups and sparse-attention (wG+SA) outperforms a vanilla encoder without groups and full-attention (woG+FA) by 1-2\% on the video model and 2-3\% for the dialog model.

Across columns, performance changes are minor.
However, when using wG+SA at the episode-level, gated attention fusion with a shot transformer ($\boxplus$) improves results
over Avg and Max pooling by 1\%.
For dialog-only, though wCLS outperforms Avg and Max by 0.7\%, we adopt Avg pooling for its effective performance in a multimodal setup.

\input{sections/tables/mm_ablations}

\paragraph{\modelname{} ablations}
are presented in \cref{tab:mm_ablations}.
Rows 1 and 2 highlight the best video-only and dialog-only models (from \cref{tab:vid_dia_abl}).
We report mean $\pm$ std dev on the val set.
Std dev is found to be high due to variation across multiple folds;
but low across random seeds.
Results for joint prediction of video shot and utterance importance are shown in rows 3-6.
Our proposed approach in row 6
performs best for both modalities, outperforming rows 3-5.

\input{sections/tables/sota_ablations}

\paragraph{SoTA comparison.}
We compare against SoTA methods:
(i)~video-only (PGLSUM~\cite{pglsum}, MSVA~\cite{msva}),
(ii)~dialog-only (PreSumm~\cite{presumm}), and
(iii)~multimodal (A2Summ~\cite{he2023a2summ}) in \cref{tab:sota_abl}.
While none of the above methods are built for processing 40 minutes of video, we make modifications to them to make them comparable to our work (details in the supplement).
On both the validation and the test set, \modelname{} outperforms all other baselines in both modalities.

\begin{figure*}
\centering
\includegraphics[width=0.99\linewidth]{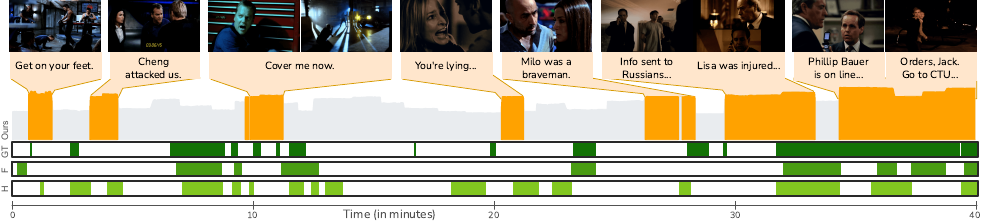}
\vspace{-1mm}
\caption{\modelname{} predictions on S06E22 of \tf{} (test set).
``Ours'' filled-plot illustrates the importance score profile over time, where {\color{yorange} orange patches} indicate story segments selected for summarization.
Annotations are shown below:
ground-truth ({\color{g1} GT}),
fandom ({\color{g2} F}), and
human annotated ({\color{g3} H}).
The story:
Amid the high-stakes sequence depicted in the selected groups 1-3, Zhou Yong's team captures Josh Bauer, leading to a firefight with Jack Bauer, who seeks Josh's location.
Negotiations with Phillip Bauer over Josh's return for a vital circuit board escalate global tensions between Russia and the USA.
Simultaneously, Mike Doyle defies Jack's wishes and departs with Josh by helicopter (segment 7).
Parallely, Lisa, backed by Tom Lennox, confronts a Russian agent, leading to her injury (4, 6).
Morris attempts to console Nadia for Milo's loss at CTU in 5.
Escalating global tensions and the imminent showdown mark the episode.}
\vspace{-2mm}
\label{fig:qualitative_result}
\end{figure*}

\subsection{Analysis and Discussion}
\label{subsec:analysis}

\input{sections/tables/benchmark_abl}

\paragraph{Video summarization benchmarks.}
We evaluate \modelname{} on SumMe~\cite{summe} and TVSum~\cite{tvsum}.
However, both datasets are small (25 and 50 videos) and have short duration videos (few minutes).
As splits and metrics are not comparable across previous works, we re-ran the baselines.

While MSVA uses three feature sets: i3d-rgb, i3d-flow~\cite{i3d} and GoogleNet~\cite{googlenet} with intermediate fusion,
PGLSUM uses GoogleNet and captures local and global features.
In contrast, A2Summ~\cite{he2023a2summ} aligns cross-modal information using dual-contrastive loss between video (GoogleNet features) and text (captions generated using  GPT-2~\cite{Radford2019LanguageMAgpt2}, embedded by RoBERTa~\cite{roberta} at frame level).

Similar to MSVA, we fuse all 3 features.
Even though \modelname{} is built for long videos (group blocks, sparse attention), \cref{tab:bench-abl} shows that we achieve SoTA on SumMe.
The drop in performance on TVSum may be due to video diversity (documentaries, how-to videos, \etc).

\input{sections/tables/split}

\paragraph{Generalization to a new season/TV series.}
\cref{tab:split_table} shows results in two different setups.
In \xseason, we see the impact of evaluating on unseen seasons (in a 7-fold cross-val-test).
While \modelname{} outperforms baselines, it is interesting that most methods show comparable performance across \intracvt{} and \xseason{} setups (see~\cref{tab:split_table,tab:sota_abl}).

In the \xseries{} setting, we train our model on \tf{} and evaluate on \pb.
Although both series are crime thrillers, there are significant visual and editing differences between the two shows.
Our approach obtains good scores on video summarization, and is a close second on dialog.

\input{sections/tables/reliability}
\input{sections/tables/fandom_human}

\paragraph{Label consistency.}
As suggested by~\cite{summe,tvsum}, label consistency is crucial to evaluate summarization methods.
We assess \dataname{} using Cronbach's $\alpha$, pairwise $F_1$-measure, and another agreement score: Fleiss' $\kappa$.

We obtain three sets of labels for 17 episodes of \tf{} (details in \cref{sup_sec:lab_rel}).
(i)~GT: obtained from matching recaps;
(ii)~F: maps plot events from a \tf{} fan site%
\footnote{\url{https://24.fandom.com/wiki/Day_6:_4:00am-5:00am} talks about the key story events in S06E22 in a \emph{Previously on 24} section (see \cref{fig:qualitative_result}).} to videos;
(iii)~H: human response for a summary.
Our labels have superior consistency compared to SumMe~\cite{summe} and TVSum~\cite{tvsum} (see \cref{tab:reliability}), indicating that identifying key story events in a TV episode is less subjective than scoring importance for generic Youtube videos.
\cref{tab:fh} shows the results for the baselines and \modelname{} on the two other labels F and H.
Our model predictions are better aligned with both labels.

\paragraph{Qualitative analysis.}
We show the model's predictions and compare against all three labels (GT, F, and H) for one episode in \cref{fig:qualitative_result}.
Our model identifies many important story segments that are also part of the annotations.

%% file: sections/tables/vid_dia_ablations.tex
\begin{table}[t]
\centering
\tabcolsep=0.09cm
\begin{tabular}{l ccccc ccc}
\toprule
\multirow{2}{*}{} &
    \multicolumn{5}{c}{Video-only} & 
    \multicolumn{3}{c}{Dialog-only} \\
\cmidrule(lr){2-6} \cmidrule(lr){7-9} 
      & Avg & Max
      & Cat & Tok
      & $\boxplus$
      & Max & Avg & wCLS \\ 
\midrule
MLP &
  42.3 &
  42.3 &
  42.3 &
  42.4 &
  42.5 &
  35.7 &
  35.7 &
  35.8 \\
woG + FA &
  51.8 &
  51.9 &
  51.1 &
  51.6 &
  52.0 &
  44.5 &
  44.5 &
  44.6 \\
wG + SA &
  52.4 &
  52.5 &
  53.3 &
  53.3 &
  \textbf{53.4} &
  46.5 &
  \emph{46.5} &
  \textbf{47.2} \\
\bottomrule
\end{tabular}
\vspace{-2mm}
\caption{
\textbf{Rows} demonstrate methods for capturing episode-level interactions.
In an \underline{MLP}, tokens are independent.
\underline{woG+FA} is a Transformer encoder that captures full-attention over the entire episode without grouping; and
\underline{wG+SA} uses the proposed architecture with local story groups and sparse-attention.
\textbf{Columns} describe the \emph{aggregation} method used to combine frame (or token) level features into shot (or utterance) representation.
C1 and C7 use average pooling.
C2 and C6 use max pooling.
C3-C5 are variants of $\STRM$:
C3 concatenates backbone features of each frame,
C4 uses backbone features as separate tokens, and
C5 uses proposed $\boxplus$ attention fusion.
C8 uses the CLS token for dialog.
Chosen: $\boxplus$ for shot, and average pooling for utterance representation.
}
\vspace{-4mm}
\label{tab:vid_dia_abl}
\end{table}

%% file: sections/tables/mm_ablations.tex
\begin{table}[t]
\centering
\small
\tabcolsep=0.12cm
\begin{tabular}{llcccc}
\toprule
 & Model & AttnMask & GToken & Video AP & Dialog AP \\ \midrule
1 & Video-only & SA & \dingcheck{} & 53.4 \scriptsize{$\pm$ 3.9} & - \\
2 & Dialog-only & SA & \dingcheck{} & - & 47.2 \scriptsize{$\pm$ 3.9} \\ \midrule
3 & \multirow{4}{*}{\modelname{}} & FA & - & 51.8 \scriptsize{$\pm$ 3.6} & 43.8 \scriptsize{$\pm$ 4.7} \\
4 &  & FA & \dingcheck{} & 51.9 \scriptsize{$\pm$ 3.7} & 44.0 \scriptsize{$\pm$ 4.6} \\
5 &  & SA & - & 53.9 \scriptsize{$\pm$ 3.4} & 48.8 \scriptsize{$\pm$ 4.6} \\
6 &  & SA & \dingcheck{} & \textbf{54.2} \scriptsize{$\pm$ 3.3} & \textbf{49.0} \scriptsize{$\pm$ 4.9} \\ \bottomrule
\end{tabular}
\vspace{-2mm}
\caption{
\modelname{} ablations.
The \emph{AttnMask} column indicates if the self-attention mask is applied over the full episode (FA) or uses sparse block diagonal structure of story groups (SA).
The \emph{GToken} indicates whether the group token is absent (-) or present (\dingcheck{}).
R6 is our final chosen model for subsequent experiments.
}
\vspace{-1mm}
\label{tab:mm_ablations}
\end{table}

%% file: sections/tables/sota_ablations.tex
\begin{table}[t]
\centering
\small
\tabcolsep=0.08cm
\begin{tabular}{l cc cc}
\toprule
\multirow{2}{*}{Model} & \multicolumn{2}{c}{Val}                    & \multicolumn{2}{c}{Test}                   \\
\cmidrule(lr){2-3} \cmidrule(lr){4-5} 
                             & Video AP              & Dialog AP               & Video AP              & Dialog AP               \\ \midrule
PGLSUM~\cite{pglsum}        & 48.8 \scriptsize{$\pm$ 3.3}          & -                    & 47.1 \scriptsize{$\pm$ 2.4}          & -                    \\
MSVA~\cite{msva}            & 47.3 \scriptsize{$\pm$ 3.8}          & -                    & 45.5 \scriptsize{$\pm$ 1.2}          & -                    \\
Video-Only                   & 53.4 \scriptsize{$\pm$ 3.9} & -                    & \textbf{50.6} \scriptsize{$\pm$ 3.6} & -                    \\ \midrule
PreSumm~\cite{presumm}      & -                   & 43.1 \scriptsize{$\pm$ 3.3} & -                   & 41.6 \scriptsize{$\pm$ 2.0} \\
Dialog-Only                  & -                   & 47.2 \scriptsize{$\pm$ 3.9}           & -                   & 43.4 \scriptsize{$\pm$ 2.8}           \\ \midrule
A2Summ~\cite{he2023a2summ}  & 35.1 \scriptsize{$\pm$ 1.8}          & 33.2 \scriptsize{$\pm$ 2.8}           & 33.8 \scriptsize{$\pm$ 1.7}          & 31.6 \scriptsize{$\pm$ 2.2}           \\
\modelname{} (Ours) & \textbf{54.2} \scriptsize{$\pm$ 3.3} & \textbf{49.0} \scriptsize{$\pm$ 4.9} & \emph{50.1} \scriptsize{$\pm$ 2.8} & \textbf{46.0} \scriptsize{$\pm$ 2.1} \\ \bottomrule
\end{tabular}
\vspace{-2mm}
\caption{
Comparison against SoTA video-only, text-only, and multimodal summarization models.
Our approach outperforms previous work by a significant margin.
}
\vspace{-3mm}
\label{tab:sota_abl}
\end{table}

%% file: sections/tables/benchmark_abl.tex
\begin{table}[t]
\centering
\small
\tabcolsep=0.18cm
\begin{tabular}{lcccccc}
\toprule
\multicolumn{1}{l}{\multirow{2}{*}{Model}} & \multicolumn{3}{c}{SumMe} & \multicolumn{3}{c}{TVSum} \\ \cmidrule(lr){2-4} \cmidrule(lr){5-7} 
\multicolumn{1}{l}{} & F1 & SP & KT & F1 & SP & KT \\ \midrule
MSVA~\cite{msva} & 52.4 & 12.3 & 9.2 & \emph{63.9} & 32.1 & 22.0 \\
PGLSUM~\cite{pglsum} & 56.2 & 17.3 & 12.7 & \emph{63.9} & \textbf{40.5} & \textbf{28.2} \\
A2Summ~\cite{he2023a2summ} & 54.0 & 3.5 & 2.8 & 62.9 & 25.2 & 17.1 \\
\midrule
\modelname{} (Ours) & \textbf{57.5} & \textbf{23.8} & \textbf{17.6} & \textbf{64.0} & 26.7 & 18.2 \\ \bottomrule
\end{tabular}
\vspace{-2mm}
\caption{Comparison with SoTA methods on the SumMe~\cite{summe} and TVSum~\cite{tvsum} benchmark datasets.
Metrics are suggested by Otani~\etal~\cite{otani}: F1, Kendall’s $\tau$ (KT), and Spearman’s $\rho$ (SP).}
\label{tab:bench-abl}
\vspace{-3mm}
\end{table}

%% file: sections/tables/split.tex
\begin{table}[t]
\centering
\small
\tabcolsep=0.12cm
\begin{tabular}{l l cc cc}
\toprule
& & \multicolumn{2}{c}{\xseason{} (\emph{24})} & \multicolumn{2}{c}{\xseries{} (\emph{PB})}  \\ 
\cmidrule(lr){3-4} \cmidrule(lr){5-6}
& \multirow{-2}{*}{Model} & Video & Dialog & Video & Dialog                         \\
\midrule
1 & MSVA~\cite{msva}     & 46.7 \scriptsize{$\pm$ 2.7}  & -            & 32.7 & - \\
2 & PGLSUM~\cite{pglsum} & 47.1 \scriptsize{$\pm$ 2.4}  & -            & 34.5 & -                           \\
3 & PreSumm~\cite{presumm} & -           & 41.3 \scriptsize{$\pm$ 3.2}  & -    & \textbf{38.3}                        \\
4 & A2Summ~\cite{he2023a2summ} & 33.5 \scriptsize{$\pm$ 3.2}  & 31.7 \scriptsize{$\pm$ 2.9}   & 20.2 & 19.0 \\ \midrule
5 & \modelname{} (Ours) & \textbf{51.0} \scriptsize{$\pm$ 4.6} & \textbf{46.0} \scriptsize{$\pm$ 5.5} & \textbf{36.7} & \textit{35.7} \\ \bottomrule
\end{tabular}
\vspace{-2mm}
\caption{
We evaluate our model's generalization across seasons within \emph{24} and across TV shows (\emph{24} $\rightarrow$  \emph{Prison Break}).
R5 showcases superiority of our methods compared to SoTA.
In \xseries, the random baseline achieves 21.3 and 19.1 for video and dialog.
}
\vspace{-4mm}
\label{tab:split_table}
\end{table}

%% file: sections/tables/reliability.tex
\begin{table}[t]
\centering
\small
\tabcolsep=0.1cm
\begin{tabular}{lcccccc}
\toprule
\multirow{2}{*}{Dataset} & \multicolumn{2}{c}{Cronbach $\alpha$} & \multicolumn{2}{c}{Pairwise $F_1$} & \multicolumn{2}{c}{Fleiss' $\kappa$} \\
\cmidrule(lr){2-3} \cmidrule(lr){4-5} \cmidrule(lr){6-7} 
                & Video  & Dialog  & Video  & Dialog  & Video  & Dialog  \\ \midrule
SumMe           & 0.88 & -    & 0.31 & -    & 0.21    & -    \\
TVSum           & 0.98 & -    & 0.36 & -    & 0.15    & -    \\
PlotSnap (Ours) & 0.91 & 0.93 & 0.59 & 0.60 & 0.38 & 0.39 \\ \bottomrule
\end{tabular}
\vspace{-2mm}
\caption{Label consistency across datasets.
}
\label{tab:reliability}
\vspace{-2mm}
\end{table}

%% file: sections/tables/fandom_human.tex
\begin{table}[t]
\centering
\small
\tabcolsep=0.1cm
\begin{tabular}{l cc cc}
\toprule
\multirow{2}{*}{Methods} & \multicolumn{2}{c}{Fandom (F)}    & \multicolumn{2}{c}{Human (H)}     \\ \cmidrule(lr){2-3} \cmidrule(lr){4-5}
                                & Video AP        & Dialog AP        & Video AP        & Dialog AP        \\ \midrule
GT          & 64.1 & 63.8 & 44.8 & 42.2 \\
\midrule
PGLSUM      & 43.0 & -  &  47.6  & -  \\
MSVA        & 34.3 & -  &  42.2 & - \\
PreSumm     & - & 43.6 & - & 46.7 \\
A2Summ      & 28.7 & 29.7 & 41.0 &  41.1 \\ \midrule
\modelname{} & \textbf{44.5} & \textbf{45.5} &  \textbf{48.7} & \textbf{50.9} \\ \bottomrule
\end{tabular}
\vspace{-2mm}
\caption{
Results on labels from \emph{24} fan site (F) and human-annotated story summaries (H) averaged over 17 episodes of \emph{24}.}
\label{tab:fh}
\vspace{-4mm}
\end{table}

%% file: sections/06_conclusion.tex
\section{Conclusion}
\label{sec:conclusion}

Our work pioneered the use of TV episode recaps for story understanding.
We proposed \dataname{}, a dataset of two TV shows with high-quality recaps, leveraging them for story summarization labels, while showing high consistency across labeling approaches.
We introduced \modelname{}, a hierarchical summarization approach that captures and compresses shots and dialog, and enables cross-modal interactions across the entire episode, trainable on a single GPU of \SI{12}{\giga\byte}.
We performed thorough ablations,
established SoTA performance, and demonstrated transfer across seasons, other series, and even movie genres.
For reproducibility and encouraging future work, we will release the code and share the dataset, as keyframes, features, and labels.

\paragraph{Limitations and future work.}
While our current work focuses on recaps obtained from a limited genre and two TV series, we believe the approach should be scalable to additional genres and datasets.
Early experiments in evaluating our model on condensed movies (CMD)~\cite{bain2020condensed} show limited improvements.
Our approach to story summarization does not explicitly model the presence of characters (\eg~via person and face tracks and their emotions~\cite{emotx}) which are central to any story and this can be an important direction for future work.
Additional discussions are provided in the supplementary material.

{\small
\paragraph{Acknowledgments.}
We thank the Bank of Baroda for partial travel support, and
IIIT-H's faculty seed grant and Adobe Research India for funding.
Special thanks to Varun Gupta for assisting with experiments
and the Katha-AI group members for user studies.
}

%% file: sections/A_IntroSupp.tex
\section*{Appendix}

We provide additional information and experimental results to complement the main paper submission.
In this work, we propose a dataset \emph{\dataname{}} consisting of two crime thriller TV shows with rich recaps and a new hierarchical model \emph{\modelname{}} that selects important local story groups to form a representative story summary.
We present details of \dataname{} in \cref{sup_sec:dataset}.
\cref{sup_sec:expt_res} extends on additional experimentation on \dataname{} as well as SumMe~\cite{summe} and TVSum~\cite{tvsum} followed by an extensive qualitative study.
Finally, we conclude this supplement with a discussion of future research directions in \cref{sup_sec:limitations}.

%% file: sections/A_Matching.tex
\subsection{Shot Matching}
\label{sup_sec:matching}

We propose a novel shot-matching algorithm whose working principle involves frame-level similarities to obtain matches.

\begin{figure}[!htb]
\centering
\includegraphics[width=\linewidth]{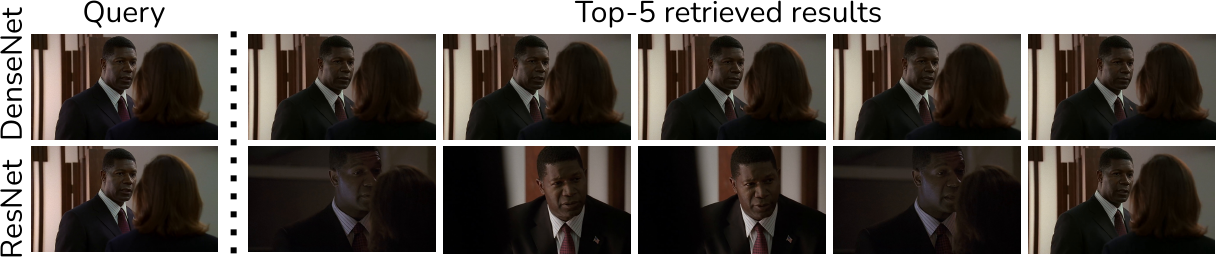}
\vspace{-4mm}
\caption{Retrieval results for Recap from Episode Frames with DenseNet (Top) v/s ResNet (bottom). We observe qualitatively that DenseNet is able to match to the correct frames from the episode more often.}
\label{fig:resVdense}
\vspace{-2mm}
\end{figure}

First, we compute frame-level embeddings using DenseNet169~\cite{densenet}, which were found to work better than models such as ResNet pre-trained on ImageNet~\cite{imagenet, He2015DeepRL_resnet} based on a qualitative analysis.
An example is shown in \cref{fig:resVdense} where we see higher similarity between retrieved episode frames and the query frame from the recap.

Second, we discuss how these embeddings are used to obtain matches is detailed below.

\begin{figure*}[t]
\centering
\includegraphics[width=0.77\linewidth]{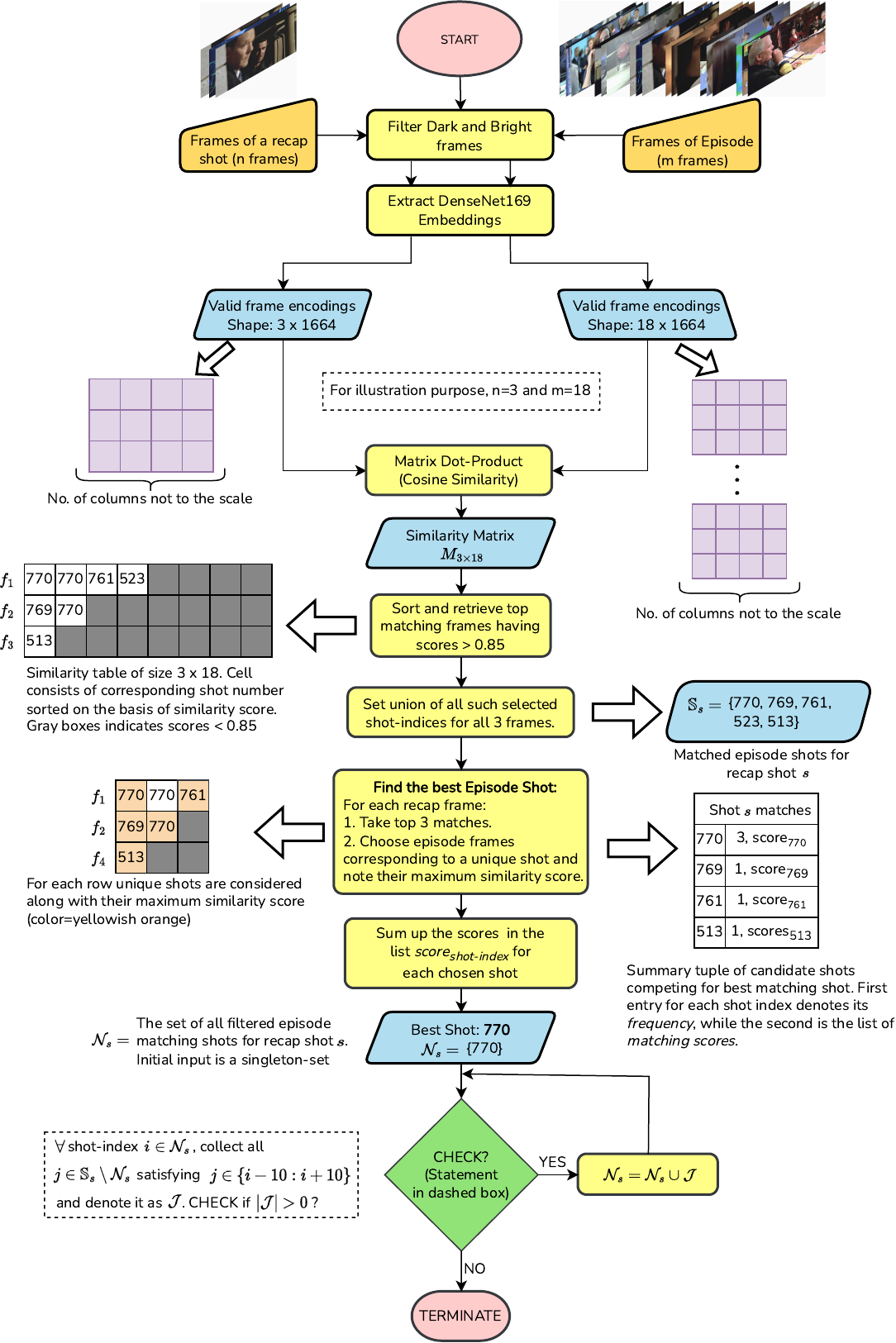}
\caption{Flowchart for identifying shots from the episode that appear in a recap and can be used as weak labels for story summarization.
The process involves identifying the list of high-scoring matching frames, indexing the shots, and then preventing spurious matches by looking for high-scoring matches within a bounded duration.
The flowchart presents an example of the process used to identify the set of shots $\MN_s$ from the episode that match to the recap shot.}
\label{fig:flow_chart}
\end{figure*}

\paragraph{Matching.}
For a given recap shot $s$ in $\MR_{n+1}$ we compare it against multiple frames in the
episode $\ME_n$, and compute a matrix dot-product with appropriate normalization (\emph{cosine similarity}) between respective frame representations of the recap and episode as illustrated in the toy-example of \cref{fig:flow_chart}.
We remove very dark or very bright frames, typical in poor lighting conditions or glares, to avoid spurious matches and noisy labels.

Next, we choose a high threshold to identify matching frames ($0.85$ in our case after analysis) and fetch all the top matching frames along with their shot indices (the shot where the frame is sourced from).
We compute a set union over all matched shots obtained by scoring similarities between the recap frame of shot $s$ and denote this set as $\mathbb{S}_{s}$.
In the example, we match 3 frames of a recap shot and identify several episode shots shown in the blue box with $\mathbb{S}_s$.

\begin{figure*}
\centering
\includegraphics[width=\textwidth]{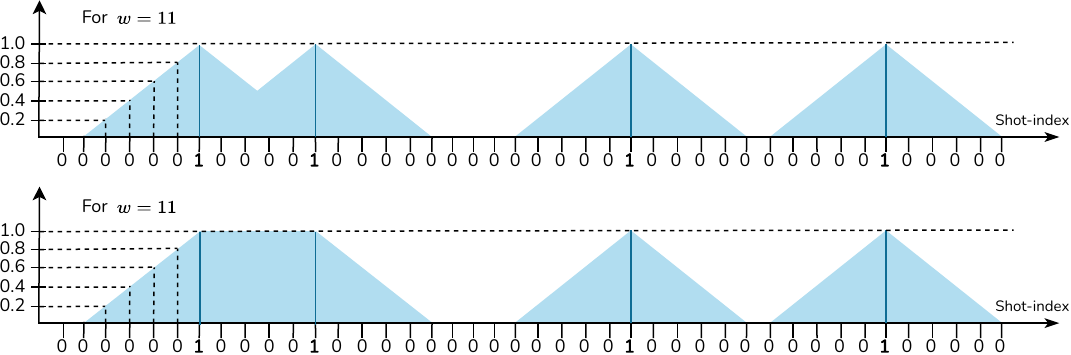}
\caption{Triangle smoothing process.
Here \emph{x-axis} denotes binary labels derived from the \emph{shot matching} process, while \emph{y-axis} shows soft \emph{importance scores} used to train our model.
\textbf{Top:} The triangular filter is applied at each shot selected from the matching process.
Scores of shots falling within the window are updated. 
\textbf{Bottom:} In the second step, we add shot importance derived potentially from multiple overlapping triangle filters.
This typically happens when episode shots in close proximity are matched to the recap.}
\label{fig:triangle_smooth}
\end{figure*}

\paragraph{Weeding out spurious shot matches.}
The set $\mathbb{S}_{s}$ may contain shots beyond a typical shot thread pattern due to spurious matches.
These need to be removed to prevent wrong importance scores from being obtained from the recap.
To do this, we first find the best matching shot in the episode.
We observe that taking top three matched frames for every recap frame results in strong matches.
Subsequently, we pick the maximum similarity score for each unique shot matched to a recap frame.
This allows us to accumulate the score for an episode shot if multiple frames of the episode shot match with frames of the recap shot.
The shot that scores the highest (after summing up the scores) is considered the best-matched episode shot for recap shot $s$.

Next, we choose a window size of 21 (10 on either side) and include all shots in $\mathbb{S}_{s}$ that fall within this window to a new matched set, $\MN_{s}$.
This is motivated by the typical duration of a scene in a movie or TV episode (40-60 seconds) and an average shot duration of 2-3 seconds.
We repeat this process until no more shots are added to the set $\MN_s$ and discard the rest in $\mathbb{S}_{s}$.
Thus, for a given shot from the recap, we obtain all matching shots in the episode that are localized to a certain region of high-scoring similarity (see Fig.~\ref{fig:flow_chart}).
We repeat this process for all frames and shots from the recap.

%% file: sections/B_Smoothing.tex
\subsection{Label Smoothing}
\label{sup_sec:smoothing}

The intuition behind extending the recap matched shots obtained in the previous section is to include chunks of the sub-story that are important to the storyline.
While a short recap (intended to bring back memories) only selects a few shots, a story summary should present the larger sub-story.
Selecting only one shot in a thread~\cite{tapaswi2014storygraphs} also adversely affects model training due to conflicting signals, as multiple shots with similar appearance can have opposite labels.
Label smoothing solves both these issues.

\paragraph{Triangle Smoother.}
We hypothesize that the importance of shots neighboring a matched shot are usually quite high and use a simple \emph{triangle smoother} to re-label the importance of shots.
In particular, we slide a window of size $w$ centering at positive labels and set the importance of neighboring shots according to height of the triangle.
The above process is illustrated in~\cref{fig:triangle_smooth} as the first step.
In the second step, we add and clip the soft labels of strongly overlapping regions to prevent any score from going higher than 1.
For the sake of simplicity, we used \emph{triangle} smoothing, however, one could also use other filters.

We choose $w = 17$ by analyzing the spread of the shots and their importance scores and comparing them against a few episodes for which we manually annotated the story summaries.

\paragraph{Dialog labels.}
The above smoothing procedure generates soft labels for video shots.
For dialog utterances, we simply import the score of the shot that encompasses the mid-timestamp of the dialog.
The key assumption here is that the dialog utterance associated with the matched shot is also important.

%% file: sections/G_Misc.tex
\subsection{Data Splits}
\label{sup_sec:splits}
For evaluation of our approach, \dataname{} is split into 4 types of splits as described below:
\begin{enumerate}[leftmargin=*]
\item \intracvt{} (Intra-Season 5 fold cross-val-test) represents 5 different non-overlapping splits from 7 seasons of \tf{} (Season 2 to 8).
\emph{Intra-season} means episodes from each season are present in the train/val/test splits.
For example, split-1 uses 5 episodes from the end of each season for val and test (2, 3 respectively). Likewise, Split-5 (the fifth fold) uses episodes from the beginning of the season in val/test.
We observe that Split-5 is harder.
\item \xseason{} (Cross-Season 7 fold cross-val-test) involves 7 non-overlapping splits with one season entirely kept for testing (from Season 2 to 8), while the train and val use 18 and 5 episodes respectively from each season.
This strategy is used to test for the generalization of our model on different seasons of the same series, \tf.
\item \xseries{} (Cross-Series) split includes 8 seasons (seasons 2 to 9) from \tf{}in train/val (19/4) and 2 seasons (seasons 2 and 3) from \emph{PB} for the test. This split is designed to check the effectiveness of our model across  TV series.
\item \emph{Multiple labels split} consists of a single non-overlapping train/val/test split with 126/18/17 episodes from \tf, respectively.
We collect story summary annotations from 3 sources for the test set (recap: GT, fan website: F, and human annotation: H).
This split is used for comparison against labels from \emph{Fandom} or \emph{Human} annotations.
\end{enumerate}

\subsection{Label Consistency}
\label{sup_sec:lab_rel}
As discussed in Tab.~8 (of the main paper), we show the agreement between story summary labels obtained from different sources.
Consistency is evaluated on 3 kinds of labels: \emph{From recap} (GT), \emph{Fandom} (F) and \emph{Human} (H).
We assess the consistency of labels via Cronbach's $\alpha$, Pairwise $F_1$, and Fleiss' $\kappa$ statistics.

\cref{tab:reliability_plus} expands on the individual scores obtained for each of the 17 episodes.
We observe that Cronbach's $\alpha$ is consistently high, while Fleiss's $\kappa$ varies typically between 0.2-0.5 indicating fair to moderate agreement.

\input{sections/tables/supplementary/reliability_plus}

%% file: sections/tables/supplementary/reliability_plus.tex
\begin{table}[t]
\centering
\small
\tabcolsep=0.14cm
\begin{tabular}{cc ccc ccc}
\toprule
\multirow{2}{*}{season} & \multirow{2}{*}{episode} & \multicolumn{3}{c}{Video} & \multicolumn{3}{c}{Dialog} \\ \cmidrule(lr){3-5} \cmidrule(lr){6-8} 
 &  & C$\alpha$ & P$F_1$ & F$\kappa$ & C$\alpha$ & P$F_1$ & F$\kappa$ \\ \midrule
S02 & E21 & 0.978 & 0.51 & 0.337 & 0.964 & 0.51 & 0.335 \\
S02 & E23 & 0.937 & 0.407 & 0.135 & 0.966 & 0.472 & 0.182 \\
S03 & E20 & 0.959 & 0.684 & 0.595 & 0.939 & 0.715 & 0.629 \\
S03 & E22 & 0.907 & 0.568 & 0.422 & 0.861 & 0.553 & 0.409 \\
S04 & E20 & 0.954 & 0.671 & 0.47 & 0.949 & 0.716 & 0.568 \\
S04 & E21 & 0.981 & 0.618 & 0.405 & 0.983 & 0.504 & 0.286 \\
S05 & E21 & 0.886 & 0.521 & 0.27 & 0.872 & 0.497 & 0.249 \\
S05 & E22 & 0.994 & 0.573 & 0.282 & 0.991 & 0.611 & 0.334 \\
S06 & E20 & 0.986 & 0.639 & 0.432 & 0.975 & 0.689 & 0.534 \\
S06 & E21 & 0.979 & 0.645 & 0.437 & 0.96 & 0.618 & 0.439 \\
S06 & E22 & 0.965 & 0.557 & 0.297 & 0.929 & 0.632 & 0.406 \\
S06 & E23 & 0.981 & 0.496 & 0.263 & 0.986 & 0.475 & 0.206 \\
S07 & E20 & 0.942 & 0.684 & 0.451 & 0.968 & 0.67 & 0.382 \\
S07 & E22 & 0.849 & 0.531 & 0.349 & 0.892 & 0.54 & 0.334 \\
S07 & E23 & 0.993 & 0.525 & 0.215 & 0.988 & 0.541 & 0.235 \\
S08 & E21 & 0.233 & 0.68 & 0.527 & 0.715 & 0.667 & 0.518 \\
S08 & E22 & 0.968 & 0.685 & 0.507 & 0.948 & 0.728 & 0.538 \\
\midrule
\multicolumn{2}{c}{Average} & 0.911 & 0.588 & 0.376 & 0.934 & 0.596 & 0.387 \\
\bottomrule
\end{tabular}
\caption{Detailed overview of reliability scores for 17 episodes from \emph{24} (test set of \emph{multiple labels split}) with the last row showing the average across all episodes. \emph{Si} and \emph{Ej} stands for Season i Episode j, C$\alpha$ for Cronbach's $\alpha$, P$F_1$ for Pairwise $F_1$, and F$\kappa$ for Fleiss' $\kappa$.}
\label{tab:reliability_plus}
\vspace{-3mm}
\end{table}

%% file: sections/C0_ImplementationDetails.tex
\subsection{Implementation Details}
\label{sup_sec:impl_details}

\paragraph{Visual features.}
We first segment the episode into shots using~\cite{dfd}.
We adopt 3 specialized backbones (for their combined effective performance; shown in \cref{tab:feat_abl_sc}):
(i)~DenseNet169~\cite{densenet} pre-trained on ImageNet~\cite{imagenet}, SVHN~\cite{svhn}, and CIFAR~\cite{cifar} for object semantics;
(ii)~MViT~\cite{mvit} pre-trained on Kinetics-400~\cite{kinetics400} for action information; and
(iii)~OpenAI CLIP~\cite{clip}, pre-trained on 4M image-text pairs, for semantic information.

\paragraph{Utterance features}.
We adapt $\text{RoBERTa-large}$~\cite{roberta} originally pretrained on the reunion of five datasets: (i)~BookCorpus~\cite{moviebook-bookcorpus}, a dataset consisting of 11,038 unpublished books; (ii)~English Wikipedia~\cite{englishwiki} (excluding lists, tables and headers); (iii)~CC-News~\cite{ccnews}, a dataset containing 63 millions English news articles crawled between September 2016 and February 2019; (iv)~OpenWebText~\cite{openwebtext}, an opensource recreation of the WebText dataset used to train GPT-2; and (v)~Stories~\cite{Trinh2018ASM-stories} a dataset containing a subset of CommonCrawl data filtered to match the story-like style of Winograd schemas~\cite{winograd}.
Together these datasets weigh 160GB of text.

Given dialogs from the episode, our fine-tuning objective is to predict the important dialogs.
We extract word/token-level representations ($\bw$) from \emph{finetuned} (but frozen) $\text{RoBERTa-large}$ ($\phi_{\MU}^{\text{FT}}$) for the task of dialog story summarization.

\paragraph{Frame sampling strategy.}
We \emph{randomly} sample up to 25 frames per shot during training as a form of data augmentation.
During inference, we use \emph{uniform} sampling.
We used fourier position embeddings $\bE^\MS_j$ for indexing video frames.

\paragraph{Architecture details.}
We experiment with the number of layers for $\STRM$, $H_{\mathsf{S}}{\in}[1:3]$ and
$\ETRM$, $H_{\mathsf{E}}{\in}[1:9]$, and find $H_{\mathsf{S}}{=}1$ and $H_{\mathsf{E}}{=}6$ to work best.
Except the number of layers, $\STRM$ and $\ETRM$ have the same configuration: 8 attention heads and $D{=}128$.
Appropriate padding and masking is used to create batches.
We compare multiple local story group sizes $n_g \in \{5:30:5\}$ and find $n_g{=}20$ to work best.

\paragraph{Training details.}
Our model is trained on 4 RTX-2080 Ti GPUs for a maximum of 65 epochs with a batch size of 4 (\ie~4 entire episodes -- each GPU handling one episode).
We adopt the AdamW optimizer~\cite{adamw} with parameters:~learning rate${=}10^{-4}$, weight decay${=}10^{-3}$.
We use OneCycleLR~\cite{Smith2018SuperconvergenceVF} as learning rate scheduler with max lr${=}10^{-3}$, and multiple dropouts~\cite{dropout}: $0.1$ for projection to 128 dim inside video and utterance encoder; $0.2$ for attention layers; and $0.2$ for the classification head.
The hyperparameters are tuned for best performance on validation set.

%% file: sections/C_FeatExtract.tex
\subsection{Feature Extraction}
\label{sup_sec:feat_extraction}

Prior to feature extraction, setting an appropriate fps for every video is important to trade off between capturing all aspects of the video while keeping computational load low.
We find 8 fps to be a good balance between the two.

\paragraph{Visual Feature Backbones}
We present the details for three backbones capturing different aspects of a video.

\paragraph{DenseNet169} $\bbf^1$.
Feature extraction of salient objects/person in each frame is of utmost priority and is achieved through DenseNet169~\cite{densenet} pretrained on ImageNet~\cite{imagenet}, SVHN~\cite{svhn}, and CIFAR~\cite{cifar}.
We consider the frozen backbone without the linear classification head to obtain flattened features, $\bbf^1 \in \bbR^{1664}$. Before feeding the images, we apply a few preprocessing steps to sub-sample raw images.
\begin{enumerate}[nosep]
\item Frames are resized to $256{\times}256$ resolution along with center cropping.
\item RGB pixel values are scaled to $[0, 1]$ followed by mean and standard deviation normalization.
\end{enumerate}

We use the architecture as well as parameters from PyTorch Hub\footnote{\url{https://pytorch.org/hub/}}, version \texttt{pytorch/vision:v0.10.0}.

\paragraph{MVIT} $\bbf^2$.
Beyond objects/person, their actions too affect the importance of a shot, and hence having them serves the purpose of representing a shot from a different perspective.
For this, we use MViT~\cite{mvit} pretrained on Kinetics-400~\cite{kinetics400}.
We feed the original video with some pre-processing as explained above to obtain feature embeddings, $\bbf^2 \in \bbR^{768}$.
With a window-size${=}32$ and stride${=}16$, we extracted per-window encodings while padding zeros at the end (black-frames) to account for selecting window-size amount of frames.
\begin{enumerate}[nosep]
\label{mvit:process}
\item Frame resizing to 256${\times}$256 followed by center-cropping to 224${\times}$224 resolution.
\item Pixel scaling from \emph{8-bit} format to \emph{float} format ($[0, 1]$).
Following this, mean and standard deviation normalization is performed.
\item We chose MViT-Base $32{\times}3$ that ingests a chunk of video (32 frames) at once and produces an embedding vector.
\end{enumerate}
We import the architecture and pretrained parameters from PytorchVideo\footnote{\url{https://pytorchvideo.org/}}.

\paragraph{CLIP} $\bbf^3$.
This is a multi-modal backbone that can produce representations corresponding to matching textual descriptions.
We borrow the CLIP~\cite{clip} pre-trained model from the huggingface~\cite{huggingface} library and use their inbuilt image processor as well as feature extractor to obtain subsampled frame-level encodings, $\bbf^3 \in \bbR^{512}$.
\begin{enumerate}[nosep]
\label{clip:process}
\item Short-side is resized to 224 pixels followed by center-cropping (to $224{\times}224$).
\item Re-scaling 8-bit image to $[0, 1]$ interval.
\item Mean and standard deviation normalization.
\end{enumerate}

\paragraph{Dialog Features}
We test three different dialog features and present the details as follows.
\begin{figure}[t]
\centering
\includegraphics[width=\linewidth]{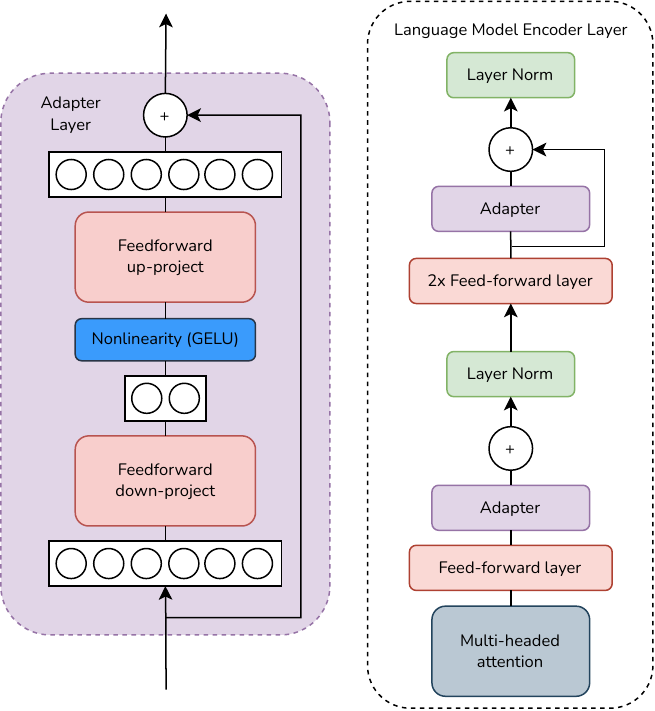}
\caption{Architecture of the adapter module~\cite{adaptor} and its integration with language model's encoder layer.
\textbf{Left}: The adapter module has fewer parameters compared to the attention and feedforward layers in a Transformer layer and consists of a bottleneck and skip-connection.
\textbf{Right}: We add the adapter module twice to each encoder layer. Once after the multi-head attention and the other is placed after the two feed-forward layers typical of a Transformer architecture.
For the Transformer \emph{decoder}, in the case of PEGASUS$_\text{LARGE}$, we add one extra adapter after cross-attention as well.
\textbf{Adaptation}: When adapting the model, the purple and green layers illustrated on the right module are trained on the downstream data and task, while the other blocks are frozen.}
\label{fig:adapter}
\end{figure}

\paragraph{Fine-tuning Language Models.}
We fine-tune language models such as PEGASUS$_\text{LARGE}$~\cite{pegasus}, RoBERTa-Large~\cite{roberta}, and MPNet-base-v2~\cite{mpnet} for our task of extractive dialog summarization.
To account for the small dataset sizes, for all fine-tuning, we use the Adapter modules~\cite{adaptor} that add only a few trainable parameters in the form of down- and up-projection layers as illustrated in Fig.~\ref{fig:adapter}.

RoBERTa-Large~\cite{roberta} and MPNet-base-v2~\cite{mpnet} are fine-tuned for utterance-level binary classification to decide whether the dialog utterance is important or not.

PEGASUS$_\text{LARGE}$ is trained originally to generate abstractive summaries.
Instead, we adapt it to generate summary dialogs.
As the number of tokens accepted by the PEGASUS$_\text{LARGE}$ model does not allow feeding all the dialog from the episode, we break it into 6 chunks.

\paragraph{Word-level embeddings.} We use \emph{last hidden-state} of the encoder to obtain contextualized word-level embeddings, $\bw \in \bbR^{1024}$ ($768$ for MPNet-base-v2).
PEGASUS$_\text{LARGE}$, being a generative model (consisting of both encoder and decoder), we keep only the encoder portion for word-level feature extraction.

%% file: sections/D_Ablations.tex
\subsection{Feature and Architecture Ablations}
\label{sup_sec:ablations}

\begin{figure*}[t]
\centering
\includegraphics[width=0.45\linewidth]{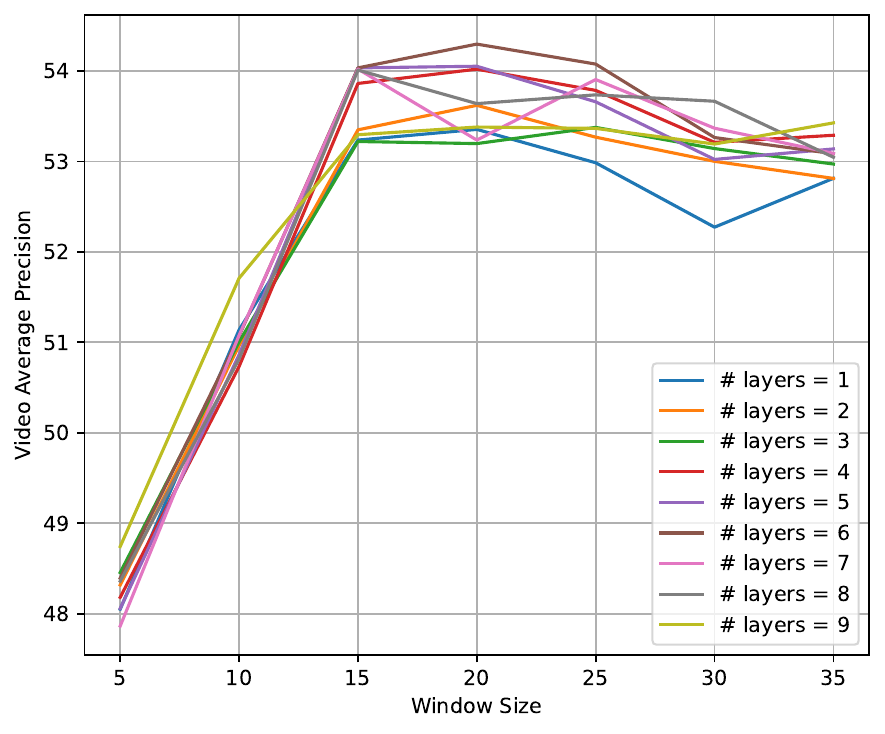}
\quad
\centering
\includegraphics[width=0.45\linewidth]{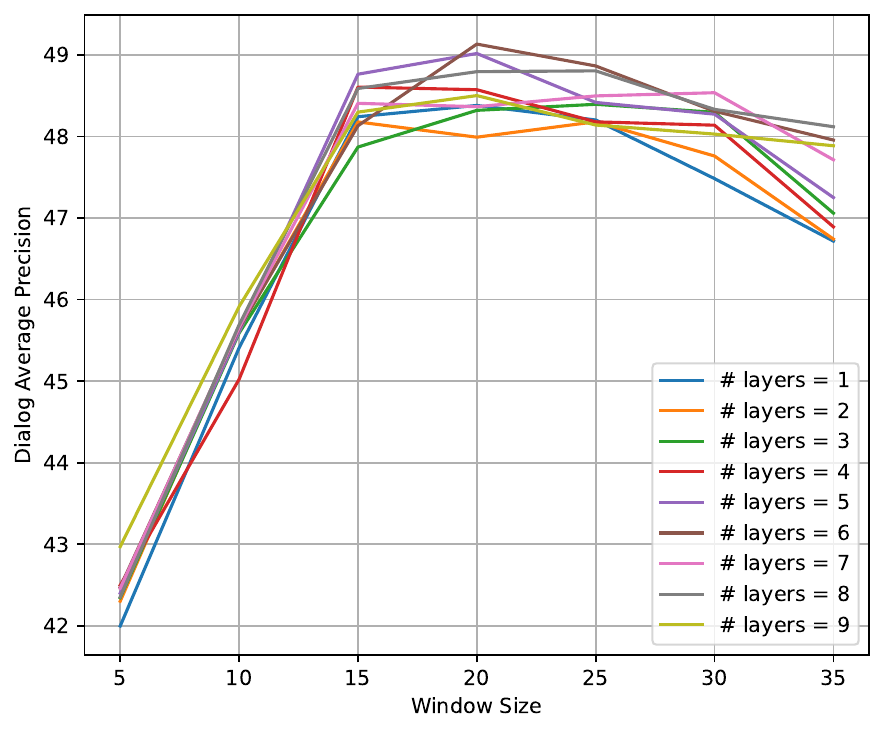}
\caption{Performance of \modelname{} for varying local story group size (also referred to as window size) and number of layers in the episode-level Transformer $\ETRM$.
\emph{y}-axis denotes the AP score for the
\textbf{Left}: Video and \textbf{Right}: Dialog.
We observe that a 6 layer model $H_\text{E} = 6$ works well together with a local story group size of $n_g = 20$.}
\label{fig:window_layer}
\end{figure*}

All experiments are run on \emph{IntraCVT} split, and we report mean and standard deviation.

\input{sections/tables/supplementary/video_feats_ablation_storyCondenser}

\paragraph{Visual features for Story summarization.}
Table~\ref{tab:feat_abl_sc} shows the results for all combinations of the three chosen visual feature backbones in a multimodal setup.
We draw two main conclusions:
(i)~The Shot Transformer encoder ($\STRM$) shows improvements when compared to simple average or max pooling.
(ii)~DMC (DenseNet + MViT + CLIP), the feature combination that uses all backbones, performs well, while MC shows on-par performance.
Importantly, the feature combination is better than using any feature alone.

\input{sections/tables/supplementary/dia_feats_ablation_storyCondenser}

\paragraph{Language backbones for Story summarization.}
Different from the previous experiment, Table~\ref{tab:dia_feats_ablation_storyCondenser} shows results for different dialog features with different word-to-utterance pooling approaches in a multimodal setting. RoBERTa-Large outperforms all other language models.
Nevertheless, the other models are not too far behind.

\paragraph{Number of $\ETRM$ layers and local story group size.}
Two important hyperparameters for our model are the number of layers $H_\text{E}$ in the $\ETRM$ and the local story group size $n_g$.
Fig.~\ref{fig:window_layer} shows the performance on video AP (left) and dialog AP (right) with a clear indication that 6 layers and a story group size of 20 shots are appropriate.

\begin{figure*}[!htb]
\centering
\includegraphics[width=\textwidth]{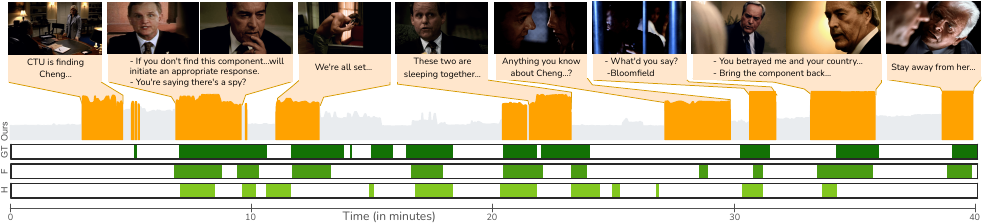}
\caption{\modelname{} predictions on S06E20 of \tf{} (test set).
``Ours'' filled-plot illustrates the importance score profile over time, where {\color{yorange} orange patches} indicate story segments selected for summarization.
Annotations are shown below:
ground-truth ({\color{g1} GT}),
fandom ({\color{g2} F}), and
human annotated ({\color{g3} H}).
We number the grouped frames representing the predicted contiguous orange chunks as shot groups (SG-n), \eg~this episode has 8 SGs.
\textbf{The story}: The White House directs CTU to locate Cheng, as depicted in SG-1, who possesses a Russian sub-circuit board that threatens national security. In SG-2, President Suvarov warns of military consequences if the Chinese agent with the circuit board isn't intercepted.
SG-3,4,7 shows how Lennox suspects a spy within the administration and uncovers Lisa's treason. President Noah Daniels instructs Lisa to bring the component back by misleading her partner, Mark Bishop. 
In SG-5,6, Jack questions Audrey about Cheng, leading to a standoff with Doyle. Audrey mentions ``Bloomfield," prompting research by Chloe. In his holding room, Heller warns Jack to stay away from Audrey due to the deadly consequences associated with him (SG-8). Intricate relationships and the imminent threat of international conflict mark the overall content of this episode.}
\label{fig:qa_1}
\end{figure*}

\begin{figure*}[!thpb]
\centering
\includegraphics[width=\textwidth]{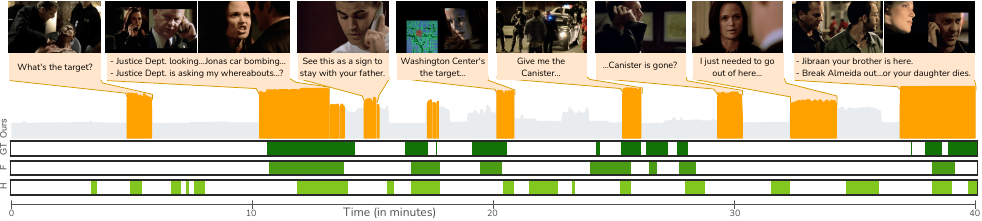}
\caption{\modelname{} predictions on S07E22 of \tf{} (test set).
``Ours'' filled-plot illustrates the importance score profile over time, where {\color{yorange} orange patches} indicate story segments selected for summarization.
Annotations are shown below:
ground-truth ({\color{g1} GT}),
fandom ({\color{g2} F}) and
human annotated ({\color{g3} H}).
We number the grouped frames representing the predicted contiguous orange chunks as shot groups (SG-n), \eg~this episode has 8 SGs.
\textbf{The story}: In a tense sequence, as shown in SG-1, Jack resorts to torture to extract information from Harbinson about the impending attack but is left empty-handed. In SG-2, following the murder of Jonas Hodges, Olivia Taylor faces scrutiny from the Justice Department. Meeting with Martin Collier, she denies transferring funds, revealing a sinister plot. Meanwhile, SG-3 shows Kim Bauer's plans are disrupted by a flight delay, leading to a strained father-daughter relationship. SG-4,5 displays how Jack, aided by Chloe O'Brian and Renee Walker, captures Tony Almeida and interrogates him about a dangerous canister, followed by Renee uncovering Jibraan's location, and a high-stakes exchange ensues at the Washington Center station. Jack detonates the canister, succumbing to its effects. As a consequence~(SG-6), Cara Bowden reports Tony's failure to Alan Wilson, adding tension to the unfolding crisis. Olivia returns to the White House, explaining her absence to Aaron Pierce (SG-7), which beautifully connects back to the SG-2. The narrative takes a dire turn as Cara blackmails Jack for the safety of Kim (SG-8), introducing a new layer of suspense and complexity to the unfolding events. \emph{The presence of SG-1,6,7 (absent in GT) clearly highlights our model's ability to complete the overall story arc.}
}
\label{fig:qa_2}
\end{figure*}

\begin{figure*}[t]
\centering
\includegraphics[width=\textwidth]{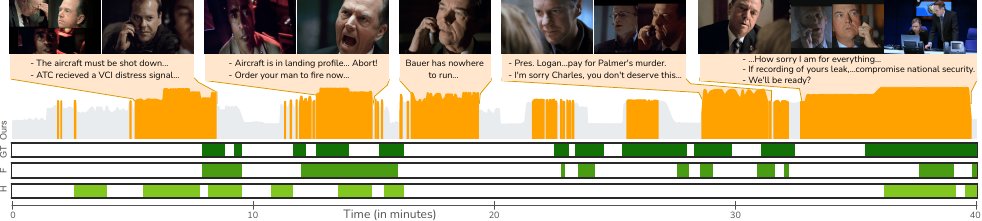}
\caption{\modelname{} predictions on S05E21 of \tf{} (test set).
``Ours'' filled-plot illustrates the importance score profile over time, where {\color{yorange} orange patches} indicate story segments selected for summarization.
Annotations are shown below:
ground-truth ({\color{g1} GT}),
fandom ({\color{g2} F}), and
human annotated ({\color{g3} H}).
We number the grouped frames representing the predicted contiguous orange chunks as shot groups (SG-n), \eg~this episode has 5 SGs.
This episode stands out due to its rapid and significant story advancements, where each sub-story holds apparent importance. 
Also, human annotations are a bit off in comparison to the ground-truth (can be verified from reliability scores shown in~\cref{tab:reliability_plus}).
Importantly, our model considers opinions from all the sources.
\textbf{The story}: In the SG-1,2, President Logan, pretending surprised, learns from Admiral Kirkland that Flight 520, now under Jack's control, is a potential threat. Despite Mike's doubts about Jack's intentions, Kirkland urges immediate action, advocating for shooting down the plane. Logan, pretending shock, reluctantly authorizes the attack. Karen alerts Jack to the order, leading to a tense situation. As the plane assumes a landing profile, Kirkland suggests calling off the strike, but Pres. Logan insists on taking it down.
Further, in SG-3, Graem criticizes Logan's decision, emphasizing the importance of capturing Jack, but Logan assures Graem of recapturing him.
Meanwhile, in SG-4, Jack, having secured incriminating evidence, vows to make Logan pay for President Palmer's assassination.
In a surprising turn, as shown in SG-5, President Logan contemplates suicide, but an unexpected call from Miles Papazian presents an alternative -- the destruction of the recording.
Encouraging Miles to act, Logan faces a critical juncture in the unfolding crisis.
Overall the entire episode sets the stage for a series of dramatic events, stressing the depth of deceit and the potential consequences for key characters.}
\label{fig:qa_3}
\end{figure*}

%% file: sections/tables/supplementary/video_feats_ablation_storyCondenser.tex
\begin{table*}[t]
\centering
\tabcolsep=0.45cm
\begin{tabular}{l l cccccc}
\toprule
\multicolumn{2}{l}{Pooling}     & Avg        & Max        & Cat        & Tok        & Stack      & $\boxplus$   \\ \midrule
\multicolumn{2}{l}{Concatenate} & \dingcheck & \dingcheck & \dingcheck & NC          & \dingcheck & \dingcheck \\ \midrule
\multirow{2}{*}{DMC} & Vid AP & 54.1 \scriptsize{$\pm$ 4.5} & 54.1 \scriptsize{$\pm$ 4.6} & \emph{54.2} \scriptsize{$\pm$ 4.1} & 54.1 \scriptsize{$\pm$ 4.0} & 53.9 \scriptsize{$\pm$ 4.7} & \textbf{54.2} \scriptsize{$\pm$ 3.3} \\
                      & Dlg AP  & 48.7 \scriptsize{$\pm$ 4.1} & 48.8 \scriptsize{$\pm$ 4.4} & 48.9 \scriptsize{$\pm$ 4.7} & \emph{49.0} \scriptsize{$\pm$ 4.8} & \emph{49.1} \scriptsize{$\pm$ 4.8} & \textbf{49.0} \scriptsize{$\pm$ 4.9} \\ \midrule
\multirow{2}{*}{DM}   & Vid AP  & 54.0 \scriptsize{$\pm$ 3.8} & 54.1 \scriptsize{$\pm$ 3.3} & 54.0 \scriptsize{$\pm$ 4.1} & 53.9 \scriptsize{$\pm$ 3.2} & 53.6 \scriptsize{$\pm$ 4.3} & 53.7 \scriptsize{$\pm$ 3.0} \\
                      & Dlg AP  & 48.6 \scriptsize{$\pm$ 4.4} & 48.8 \scriptsize{$\pm$ 4.4} & 48.9 \scriptsize{$\pm$ 4.4} & 48.6 \scriptsize{$\pm$ 3.6} & 48.4 \scriptsize{$\pm$ 4.8} & 48.6 \scriptsize{$\pm$ 4.3} \\ \midrule
\multirow{2}{*}{MC}   & Vid AP  & 53.8 \scriptsize{$\pm$ 3.6} & 53.9 \scriptsize{$\pm$ 4.0} & 53.8 \scriptsize{$\pm$ 4.5} & 54.0 \scriptsize{$\pm$ 4.9} & 53.8 \scriptsize{$\pm$ 4.2} & \textbf{54.2} \scriptsize{$\pm$ 4.3} \\
                      & Dlg AP  & 48.5 \scriptsize{$\pm$ 4.6} & 48.8 \scriptsize{$\pm$ 4.1} & 48.7 \scriptsize{$\pm$ 4.5} & \emph{49.0} \scriptsize{$\pm$ 4.1} & 48.5 \scriptsize{$\pm$ 4.7} & \textbf{49.1} \scriptsize{$\pm$ 4.2} \\ \midrule
\multirow{2}{*}{DC}   & Vid AP  & 54.0 \scriptsize{$\pm$ 4.5} & 54.1 \scriptsize{$\pm$ 4.3} & 54.1 \scriptsize{$\pm$ 4.0} & 54.0 \scriptsize{$\pm$ 4.0} & 54.1 \scriptsize{$\pm$ 3.9} & 54.1 \scriptsize{$\pm$ 4.2} \\
                      & Dlg AP  & 48.2 \scriptsize{$\pm$ 4.9} & 48.7 \scriptsize{$\pm$ 4.5} & 48.9 \scriptsize{$\pm$ 4.7} & \emph{49.0} \scriptsize{$\pm$ 4.6} &\emph{ 49.0} \scriptsize{$\pm$ 4.8} & \emph{49.0} \scriptsize{$\pm$ 4.9} \\ \midrule
\multirow{2}{*}{D}    & Vid AP  & 53.5 \scriptsize{$\pm$ 4.2} & 53.4 \scriptsize{$\pm$ 4.2} & 54.0 \scriptsize{$\pm$ 3.8} & -          & -          & -          \\
                      & Dlg AP  & 48.1 \scriptsize{$\pm$ 4.2} & 48.6 \scriptsize{$\pm$ 4.3} & 48.9 \scriptsize{$\pm$ 4.4} & -          & -          & -          \\ \midrule
\multirow{2}{*}{M}    & Vid AP  & 52.9 \scriptsize{$\pm$ 3.2} & 53.6 \scriptsize{$\pm$ 3.6} & 53.5 \scriptsize{$\pm$ 3.4} & -          & -          & -          \\
                      & Dlg AP  & 48.2 \scriptsize{$\pm$ 4.0} & 48.4 \scriptsize{$\pm$ 3.9} & 48.7 \scriptsize{$\pm$ 4.6} & -          & -          & -          \\ \midrule
\multirow{2}{*}{C}    & Vid AP  & 53.9 \scriptsize{$\pm$ 3.7} & 53.9 \scriptsize{$\pm$ 3.9} & 54.1 \scriptsize{$\pm$ 3.7} & -          & -          & -          \\
                      & Dlg AP  & 48.6 \scriptsize{$\pm$ 4.2} & 48.7 \scriptsize{$\pm$ 4.2} & 48.7 \scriptsize{$\pm$ 4.0} & -          & -          & -          \\ \bottomrule
\end{tabular}
\caption{\modelname{} ablations for different feature combination strategies, using \textbf{both video and dialog modalities}.
Feature ablations are performed over the visual modality.
Columns describe the \emph{Pooling}
approaches used to form shot-level from frame-level representations;
\emph{Concatenate} corresponds to how backbone feature are combined (concatenate \dingcheck{} or as separate tokens (NC)).
\emph{Stack} pooling is an alternative approach to $\boxplus$, where instead of condensing the aggregated (concatenated) visual features via a linear layer $\bW_{\MP} \in \bbR^{3\times 3D}$, we obtain individual feature importance score (with $\bW_{\MP} \in \bbR^{1\times D}$ followed by $\tanh$ and $\soft$).
\textbf{D}: DenseNet169,
\textbf{M}: MViT, and
\textbf{C}: CLIP.
All results are for \modelname{} that captures \emph{Episode} level interactions.
}
\label{tab:feat_abl_sc}
\end{table*}

%% file: sections/tables/supplementary/dia_feats_ablation_storyCondenser.tex
\begin{table}[t]
\centering
\small
\tabcolsep=0.09cm
\begin{tabular}{ll ccc}
\toprule
\multicolumn{2}{l}{Pooling}          & Max        & Avg        & wCLS       \\ \midrule
\multirow{2}{*}{PEGASUS$_\text{LARGE}$~\cite{pegasus}}    & Vid AP & 52.7 \scriptsize{$\pm$ 4.3} & 53.1 \scriptsize{$\pm$ 4.2} & 53.1 \scriptsize{$\pm$ 4.7} \\
                            & Dlg AP & 47.9 \scriptsize{$\pm$ 3.6} & 48.0 \scriptsize{$\pm$ 5.1} & 47.9 \scriptsize{$\pm$ 4.9} \\ \midrule
\multirow{2}{*}{MPNet-Base~\cite{mpnet}} & Vid AP & 53.2 \scriptsize{$\pm$ 3.9} & 53.1 \scriptsize{$\pm$ 3.7} & 53.5 \scriptsize{$\pm$ 3.4} \\
                            & Dlg AP & 48.0 \scriptsize{$\pm$ 4.4} & 47.2 \scriptsize{$\pm$ 3.2} & 48.6 \scriptsize{$\pm$ 5.0} \\ \midrule
\multirow{2}{*}{RoBERTa-Large~\cite{roberta}}    & Vid AP & 54.1 \scriptsize{$\pm$ 3.8} & \textbf{54.2} \scriptsize{$\pm$ 3.3} & 54.1 \scriptsize{$\pm$ 4.2} \\
                            & Dlg AP & \emph{49.0} \scriptsize{$\pm$ 4.6} & \textbf{49.0} \scriptsize{$\pm$ 4.9} & \emph{49.0} \scriptsize{$\pm$ 4.3} \\ \bottomrule
\end{tabular}
\vspace{-2mm}
\caption{\modelname{} ablations for various dialog utterance feature backbones.
Visual features are fixed to DMC.}
\label{tab:dia_feats_ablation_storyCondenser}
\vspace{-4mm}
\end{table}

%% file: sections/E_SoTA_Adapt.tex
\subsection{Adapting SoTA Approaches for~\dataname}
\label{sup_sec:adapt}

In this section, we discuss how we adapt different video- and dialog-only state-of-the-art baselines for our task.

\paragraph{MSVA~\cite{msva}}
considers frame-level features from multiple sources and applies aperture-guided attention across all such feature streams independently, followed by intermediate fusion and a linear classification head that selects frames based on predicted importance scores.
Since we are modeling at the level of the entire episode, we feed condensed shot features (after Avg or Max pooling or $\STRM$) of each backbone: 
DenseNet169 ($X_o$), MViT ($X_r$), and CLIP ($X_f$) through \emph{three} different input streams and output shot-level importance scores.

Our model is different from MSVA as MSVA treats each feature separately, while we perform early fusion and concatenate representations from multiple backbones even before obtaining a compact video shot representation.
\modelname{} is also developed for encoding and making predictions on multiple modalities, while MSVA is not.

\paragraph{PGL-SUM~\cite{pglsum}}
splits the video in small equal-sized group of frames.
Similar to our work, contextualization is performed within local multi-headed self-attention on small groups, while another is at global level using global multi-headed attention for the entire video.
Later, both are merged via addition along the feature axis and subsequently passed through an MLP classification head to obtain frame-level scores.
To adapt PGL-SUM to our work, we again think of shots as the basic unit and concatenate visual features from all streams ($\bbf^1, \bbf^2$, and $\bbf^3$), followed by pooling ($\STRM$ or Max or Avg pooling) and then PGL-SUM to generate shot-level importance scores.

PGL-SUM has some similarities to our approach as both involve local groups.
Interestingly, PGL-SUM creates groups of frames to perform summarization for short videos of a few minutes, while \modelname{} creates groups of shots and dialog utterances to generate summaries for 40 minute long episodes.
Among technical contributions, we also explore different attention mechanisms such as across the full-episode (FE) or within a local story group (SG).
Different from PGL-SUM, we introduce a story group token that allows to capture the essence of a local story group.

\paragraph{PreSumm~\cite{presumm}}
is used for text-only extractive summarization which takes word-level inputs and produces sentence-level probability scores. 
To represent each episode, the utterances are concatenated, lower-cased, and separated by $\CLS$ and $\mathsf{SEP}$ tokens into a single line input.
The PreSumm model leverages word embeddings from pre-trained BERT-base~\cite{bert} language model.
Considering the long inputs in our case, we extend the existing positional embeddings of BERT from 512 to 10000 by keeping the original embeddings and replicating the last embeddings for the remainder.
At the sentence level, the corresponding $\CLS$ token is fed into two transformer encoder layers for contextualization, followed by a small MLP and sigmoid operation to generate per-sentence scores.
The model is trained using the Adam~\cite{adam} optimizer with Binary Cross-Entropy loss.

PreSumm is very different from our work as it operates directly on tokens, while our model develops a hierarchical approach going from words to dialogs to local story groups.

\paragraph{A2Summ~\cite{he2023a2summ}} is a contemporary multimodal summarization (MSMO) baseline, primarily designed to align temporal correspondence between video and text signals through a dual-contrastive loss approach.
They also introduce a dataset, BLiSS~\cite{he2023a2summ}, comprising 13,303 pairs of livestream videos and transcripts, each with an average duration of 5 minutes, along with multimodal (VT2VT) summaries.
A2Summ exploits cross-modality correlations within and between videos through dual contrastive losses.
These include:
(a)~inter-sample contrastive loss (operates across different sample pairs within a batch, leveraging the intrinsic correlations between video-text pairs), and
(b)~an intra-sample contrastive loss (works within each sample pair, emphasizing the similarities between ground-truth video and text summaries while contrasting positive features with hard-negative features).

We adapt A2Summ for \dataname, where we work at the episodic level, by using max-pooling with an MLP for video features and average-pooling for dialog (text) features to derive shot- and utterance-level representations.
We create explicit intra-sample attention masks encouraging temporal alignment, allowing video shots to attend to their corresponding utterances and permitting video and utterance tokens to fully attend to their respective counterparts.
Considering memory constraints, we maintain a batch size of 4 (four entire episodes - inter-sample contrasting) on a single NVIDIA GeForce RTX-2080 Ti GPU.
We adopt CyclicLR~\cite{Smith2015CyclicalLR} with a maximum learning rate of $10^{-4}$ and the `triangular2' mode.
A2Summ model comprises 6 encoder layers and incorporates multiple dropout layers~\cite{dropout}:
(a)~\texttt{dropout\_video} ${=}0.1$,
(b)~\texttt{dropout\_text} ${=}0.2$,
(c)~\texttt{dropout\_attn} ${=}0.3$, and
(d)~\texttt{dropout\_fc} ${=}0.5$, while keeping rest the of the hyperparameters the same.
Training extends to a maximum of 50 epochs, with the AdamW~\cite{adamw} optimizer utilized, featuring a learning rate of $10^{-5}$ and a weight decay~\cite{Loshchilov2017DecoupledWD} ${=}0.01$.

In \dataname{}, where episodes show related content, the application of inter-episode contrastive learning negatively impacts the model's performance. 
This is due to the variability in the importance of related story segments across episodes, which depends on the specific context.

%% file: sections/F_Benchmark.tex
\subsection{Details for SumMe and TVSum}
\label{sup_sec:bench_data}
\input{sections/tables/supplementary/bench-hyper}

In this section, we will discuss how we adapted our model for SumMe~\cite{summe} and TVSum~\cite{tvsum}, some experimentation details, and corresponding evaluation metrics.

\paragraph{Adaptation.} We used our video-based model on both datasets, inheriting features from MSVA\footnote{\url{https://github.com/TIBHannover/MSVA/tree/master}}~\cite{msva}.
To capture shot-level details, we stacked 15 contiguous frame-embeddings (as previous methods utilized ground-truth labels indexed at every 15th frame), and for group-level, we used \texttt{n\_frame\_per\_seg} attribute of the dataset.
We used continuous-index-based time embeddings for shot-frames.
Essentially, we assume that a shot consists of 15 frames as SumMe and TVSum require predictions at every 15 frames.

\paragraph{Hyperparameter configuration.} We determine the configuration based on the best validation score obtained over five random splits (5-RCV).
This time our model is trained on a single NVIDIA GeForce RTX-2080 Ti GPU for a maximum of 300 epochs for SumMe and 100 epochs for TVSum, with a batch size of 1.
Additional dataset-specific hyperparameters are detailed in Table~\ref{tab:bench-hyper}.
In common, we have AdamW optimizer~\cite{adamw} with parameters:~learning rate ${=}5{\times}10^{-5}$, weight decay~\cite{Loshchilov2017DecoupledWD} ${=}10^{-3}$.
We use CyclicLR~\cite{Smith2015CyclicalLR} for learning rate scheduling with max lr ${=}5{\times}10^{-4}$ and \emph{triangular2} mode.
ReLU is used for classification head and GELU for projection and attention modules.

%% file: sections/tables/supplementary/bench-hyper.tex
\begin{table*}[t]
\centering
\begin{tabular}{l cccccccc}
\toprule
      & \texttt{amsgrad} & \texttt{d\_model} & \texttt{drop\_fc/trm/proj} & \texttt{dec\_l} & \texttt{enc\_l} & wd     & \texttt{act\_clf/mlp/trm} & \texttt{ffs} \\ \midrule
SumMe~\cite{summe} & True    & 512     & 0.5/0.2/0.2      & 3     & 1     & 0.0001 & r/g/g           & Stack  \\
TVSum~\cite{tvsum} & False   & 768     & 0.7/0.4/0.2      & 6     & 1     & 0.01   & r/g/g           & $\boxplus$  \\ \bottomrule
\end{tabular}
\caption{Hyperparameter configuration for SumMe and TVSum.
\texttt{d\_model} specify the dimension for the transformer module's internal representation, while \texttt{dec\_l} and \texttt{enc\_l} denote $\#$ of decoder and encoder layers, respectively.
Other hyperparameters include \texttt{drop\_fc/trm/proj} for dropout at classification head, attention module, and projection module, \texttt{wd} stands for weight decay parameter (used inside AdamW~\cite{adamw}), \texttt{act\_clf/mlp/trm} for activation function used in classification head, projection, and attention module, with \texttt{r} for ReLU and \texttt{g} for GELU.
Lastly, \texttt{ffs} stands for feature fusion style, with Stack and $\boxplus$ depicting the pooling strategy showed in Table~\ref{tab:feat_abl_sc}.}
\label{tab:bench-hyper}
\end{table*}

%% file: sections/H_QA.tex
\subsection{Extended Qualitative Analysis}
\label{sup_sec:qa}

In this section, we analyze 3 more episodes and compare the model's prediction against all three labels (GT, F, and H).
Recall, the labels denoted \emph{F} (Fandom) are based on summarized plot synopses from the \tf{} fan site\footnote{\url{https://24.fandom.com/wiki/Wiki_24}} that includes the key events in the story (as a text description).
Plot synopses are short textual descriptions of the key story events of an episode.
We ask annotators to use the plot synopses and tag the start-end duration for story sequences corresponding to the description.
We refer to these labels as \emph{Fandom} (F) and use them for qualitative evaluation.
While the \emph{H}-labels are annotations from a human, based on what they feel is relevant summary as per the narrative.

\paragraph{Qualitative evaluation.}
We present importance scores for three episodes: S06E20\footnote{\url{https://24.fandom.com/wiki/Day_6:_2:00am-3:00am} talks about the key story events in S06E20 in a \emph{Previously on 24} section (see \cref{fig:qa_1}).}, S07E22\footnote{\url{https://24.fandom.com/wiki/Day_7:_6:00am-7:00am} talks about the key story events in S07E22 in a \emph{Previously on 24} section (see \cref{fig:qa_2}).}, and S05E21\footnote{\url{https://24.fandom.com/wiki/Day_5:_4:00am-5:00am} talks about the key story events in S05E21 in a \emph{Previously on 24} section (see \cref{fig:qa_3}).}, in  \cref{fig:qa_1}, \cref{fig:qa_2}, and \cref{fig:qa_3}, respectively.
We observe that the model predictions are quite good and not only match the ground-truth labels (on which the model is trained) but also the fandom and human annotations.
Please refer to the figure captions for additional comments on episode-specific remarks.

%% file: sections/I_Limitations.tex
\section{Future Work}
\label{sup_sec:limitations}

Ingesting $\sim$40 minute long videos is a challenging problem.
Aligning different modalities, such as processing a 40-minute video containing 2,500 frames (at 1fps) and 4,500 dialogue words, totaling approximately 8000 tokens, may require a larger GPU memory.
And most of the L-VLMs~\cite{Li2022BLIPBL,singh2022flava,Kim2021ViLTVT} struggle to handle contexts exceeding 8,000 tokens, and even if they do, consumer-grade GPUs may lack the capacity to accommodate them.

Our approach considers coarse-grained visual information, which we demonstrate is beneficial for story-summarization.
Considering more fine-grained visual info, such as person and face tracks across frames, and their emotions, would be useful. 
Our idea of using recap-inspired story summary labels or modeling approach are not specific to thrillers and can be easily extended to other genres and shows with recaps. 
Having speaker information in dialog utterances with mapping to character faces would probably improve performance on the summarization task.
The local story groups are a proxy to scene segments of an episode.
Replacing them with actual scene segments may improve our model's performance for summarization.

Finally, further analysis and experiments are required to determine the quality of these methods~\cite{trustworthy_methods}, particularly because evaluating long video summarization using human judgment is very time-consuming.

However, we believe that this work provides a window into this challenging problem and can help facilitate further research in this area.